\documentclass{article} 
\usepackage[affil-it]{authblk}
\usepackage[hyperref]{emnlp2020}

\usepackage{amsmath,amsfonts,bm}

\def\eqref#1{equation~\ref{#1}}

\def\1{\bm{1}}

\DeclareMathAlphabet{\mathsfit}{\encodingdefault}{\sfdefault}{m}{sl}
\SetMathAlphabet{\mathsfit}{bold}{\encodingdefault}{\sfdefault}{bx}{n}

\usepackage{url}
\usepackage{graphicx}
\usepackage{subcaption}
\usepackage{float}
\usepackage{adjustbox}
\usepackage{comment}

\newcommand\mathieu[1]{\textcolor{white}{#1}}

\title{``LazImpa'': \emph{Laz}y and \emph{Impa}tient neural agents \\learn to communicate efficiently}

\author[1]{Mathieu Rita}
\author[1,2]{Rahma Chaabouni}
\author[1,2]{Emmanuel Dupoux}
\affil[1]{Cognitive Machine Learning (ENS - EHESS - PSL Research University - CNRS - INRIA)}
\affil[2]{Facebook AI Research}
\affil[ ]{\tt {mathieu.rita@polytechnique.edu,  \{rchaabouni,dpx\}@fb.com}}

\aclfinalcopy 
\begin{document}

\maketitle

\begin{abstract}
Previous work has shown that artificial neural agents naturally develop surprisingly non-efficient codes. 
This is illustrated by the fact that in a referential game involving a speaker and a listener neural networks optimizing accurate transmission over a discrete channel, the emergent messages fail to achieve an optimal length. Furthermore, frequent messages tend to be longer than infrequent ones, a pattern contrary to the Zipf Law of Abbreviation (ZLA) observed in all natural languages. Here, we show that near-optimal and ZLA-compatible messages can emerge, but only if both the speaker and the listener are modified. We hence introduce a new communication system, ``LazImpa'', where the speaker is made increasingly \emph{laz}y, i.e.,~avoids long messages, and the listener \emph{impa}tient, i.e.,~seeks to guess the intended content as soon as possible.
\end{abstract}

\section{Introduction}
\label{section:intro2}

Recent emergent-communication studies, renewed by the astonishing success of neural networks, are often motivated by a desire to develop neural network agents eventually able to verbally interact with humans \citep{havrylov:etal:2017, Lazaridou:etal:2017}. To facilitate such interaction, neural networks' emergent language should possess many natural-language-like properties. However, it has been shown that, even if these emergent languages lead to successful communication, they often do not bear core properties of natural language \citep{Kottur:etal:2017,Bouchacourt:Baroni:2018,Lazaridou:etal:2018,chaabouni:etal:2020}.

In this work, we focus on one basic property of natural language that resides on the tendency to use messages that are close to the informational optimum. This is illustrated in the Zipf's law of Abbreviation (ZLA), an empirical law that states that in natural language, the more frequent a word is, the shorter it tends to be \citep{Zipf:1949, teahan2000, sigurd2004word, strauss2007}. Crucially, ZLA is considered to be an \emph{efficient} property of our language \cite{Gibson:etal:2019}. 
Besides the obvious fact that an efficient code would be easier to process for us, it is also argued to be a core property of natural language, likely to be correlated with other fundamental aspects of human communication, such as regularity and compositionality \citep{kirby2001}. Encouraging it might hence lead to emergent languages that are also more likely to develop these other desirable properties. 

Despite the importance of such property,  \citet{chaabouni:etal:2019} showed that standard neural network agents, when trained to play a simple signaling game \citep{Lewis:1969}, develop an inefficient code, which even displays an \emph{anti-}ZLA pattern. That is, counterintuitively, more frequent inputs are coded with longer messages than less frequent ones. This inefficiency was related to  neural networks' ``innate preference'' for long messages. In this work, we aim at understanding which constraints need to be introduced on neural network agents in order to overcome  their innate preferences and communicate efficiently, showing a proper ZLA pattern.

To this end, we 
use a reconstruction game where we have two neural network agents: speaker and listener. For each input, the speaker outputs a sequence of symbols (which constitutes the message) sent to the listener. The latter needs then to predict the speaker's input based on the given message.
Also, similarly to the previous work, inputs are drawn from a power-law distribution. 

We first describe the experimental and optimization framework (see Section \ref{section:framework2}). In particular, we introduce a new communication system called `LazImpa', comprising two different constraints (a) \emph{Laz}iness on the speaker side and (b) \emph{Impa}tience on the listener side. The former constraint is inspired by the least-effort principle which is attested to be a ubiquitous pressure in human communication \citep{piantadosi2011, Zipf:1949, Kanwal2017}. 

However, if such a constraint is applied too early, the system does not learn an efficient system. We show that incrementally penalizing long messages in the cost function enables an early exploration of the message space (a kind of `babbling phase') and prevents converging to an inefficient local minimum. 

The other constraint, on the listener side, relies on the prediction mechanism, argued to be important in language comprehension \citep[e.g.,][]{federmeier2007, altmann2009}, and is achieved by allowing the listener to reconstruct the intended input as soon as possible. We also provide a two-level analytical method: first, metrics quantifying the efficiency of a code; second, a new protocol to measure its informativeness (see Section \ref{section:method2}). Applying these metrics, we demonstrate that, contrary to the standard speaker/listener agents, our new communication system `LazImpa' leads to the emergence of an efficient code. The latter follows a \emph{ZLA-like} distribution, close to natural languages (see Sections \ref{section:results:std2} and \ref{section:results:rf2}). Besides the plausibility of the introduced constraints, our new communication system is, first, task- and architecture-agnostic (requires only communicating with sequences of symbols), and second allows stable optimization of the speaker/listener. We also show how both listener and speaker constraints are fundamental to the emergence of a ZLA-like distribution, as efficient as natural language (see Section \ref{section:results:compamodels2}).

\section{Experimental framework}
\label{section:framework2}       

We explore the properties of emergent communication in the context of referential games where neural network agents, Speaker and Listener, have to cooperatively communicate in order to win the game.

Speaker network receives an input $i \in \mathcal{I}$ and generates a message $m$ of maximum length \verb+max_len+. The symbols of the message belong to a vocabulary $V=\{s_{1},s_{2},...,s_{\verb+voc_size+-1},\verb+EOS+\}$ of size \verb+voc_size+ where \verb+EOS+ is the `end of sentence' token indicating the end of Speaker's message. Listener network receives and consumes the message $m$. Based on this message, it outputs $\hat{i}$. The two agents are successful if Listener manages to guess the right input (i.e., $\hat{i}=i$).

We make two main assumptions. First inputs are drawn from $\mathcal{I}$ following a power-law distribution, where $\mathcal{I}$ is composed of 1000 one-hot vectors.

Consequently, the probability of sampling the $k^{th}$ most frequent input is: $\frac{1/k}{\sum_{j=1}^{1000}1/j}$ modelling words' distribution in natural language \cite{zipf2013} (see details in Appendix \ref{appendix:input_space}).
Second, we experiment in the main paper with $\verb+max_len+=30$ and $\verb+voc_size+=40$.\footnote{This combination makes our setting comparable to natural  languages; the latter has no upper bound on the maximum length, also a vocabulary size of  $40$ is close  to the alphabet size of the natural languages we study of mean vocabulary size equal to $41.75$. See \citet{chaabouni:etal:2019} for more details.} 
We further discuss the influence of these assumptions in Appendix. \ref{appendix:assumption_analysis} and show the robustness of our results to assumptions change.

In our analysis, we only consider the successful runs, i.e.,~the runs with a uniform accuracy strictly higher than 97\% over all possible 1000 inputs. An emergent language consists then of the input-message mapping. That is, for each input $i \in \mathcal{I}$ fed to Speaker after successful communication, we note its output $m$. 

By $\mathcal{M}$, we define the set of messages $m$ used by our agents after succeeding in the game.

\subsection{Agent architectures}
\label{section:framework:archi2}

In our experiments, we compare two communication systems:
\begin{itemize}
    \item Standard Agents: as a baseline, composed of Standard Speaker and Standard Listener;
    \item `LazImpa': composed of \emph{Laz}y Speaker and \emph{Impa}tient Listener.  
\end{itemize}

For both Speaker and Listener, we experiment with either standard or modified LSTM architectures \cite{hochreiter1997}.

\subsubsection{Standard Agents}

\paragraph{Standard Speaker.} 
Standard Speaker is a single-layer LSTM. First, Speaker's inputs $i$ are mapped by a linear layer into an initial hidden state of Speaker's LSTM cell. Then, the message $m$ is generated symbol by symbol: the current sequence is fed to the LSTM cell that outputs a new hidden state. Next, this hidden state is mapped by a linear layer followed by a softmax to a Categorical distribution over the vocabulary. During the training phase, the next symbol is sampled from this distribution. During the testing phase, the next symbol is deterministically selected by taking the argmax of the distribution.

\paragraph{Standard Listener.}

Standard Listener is also a single-layer LSTM. Once the message $m$ is generated by Speaker, it is entirely passed to Standard Listener. Standard Listener consumes the symbols one by one, until the \verb+EOS+ token is seen (the latter is included and fed to Listener). At the end, the final hidden state is mapped to a Categorical distribution $L(m)$ over the input indices (linear layer + softmax). This distribution is then used during the training to compute the loss. During the testing phase, we take the argmax of the distribution as a reconstruction candidate.

\paragraph{Standard loss $\mathcal{L}_{std}$.} For Standard Agents, we merely use the cross-entropy loss between the ground truth one-hot vector $i$ and the output Categorical distribution of Listener $L(m)$.

\subsubsection{LazImpa}

\paragraph{Lazy Speaker.}
Lazy Speaker has the same architecture as Standard Speaker. The `Laziness' comes from a cost on the length of the message $m$ directly applied to the loss.

\begin{figure}[h]
    \centering
    \includegraphics[scale=0.3]{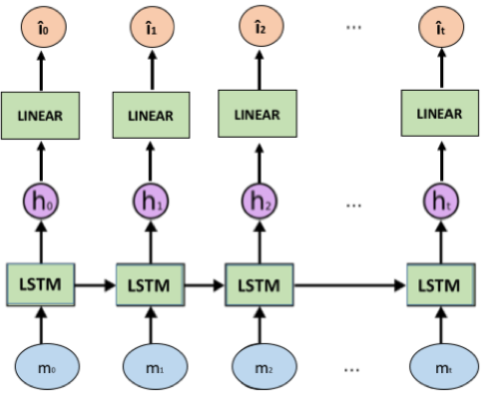}
    \caption{Impatient Listener architecture. The agent is composed of a single-layer LSTM cell and one shared linear layer followed by a softmax. It generates a prediction at each time step.
    }
    \label{fig:impatient_listener_modelN}
\end{figure}

\paragraph{Impatient Listener.}
We introduce Impatient Listener, designed to guess the intended content as soon as possible. As shown in Figure \ref{fig:impatient_listener_modelN}, Impatient Listener consists of a modified Standard Listener that, instead of guessing $i$ after consuming the entire message $m=(m_{0},...,m_{t})$, makes a prediction $\hat{i}_k$ for each symbol $m_k$.\footnote{$m_{t}$=$\texttt{EOS}$ by construction.} This modification takes advantage of the recurrent property of the LSTM, however, could be adapted to any causal sequential neural network model. 

At training, a prediction of Impatient Listener, at a position $k$, is a Categorical distribution $L(m_{:k})$, constructed using a shared single linear layer followed by a softmax (with $m_{:k}=(m_{0},...,m_{k})$). Eventually, we get a sequence of $t+1$ distributions $L(m)=(L(m_{:0}),... ,L(m_{:t}))$, one for each reading position of the message. 

At test time, we only take the argmax of the distribution generated by Listener when it reads the \verb+EOS+ token.

\paragraph{LazImpa Loss $\mathcal{L}_{laz}$.} LazImpa loss is composed of two parts that model `Impatience' ($\mathcal{L}_{laz/L}$) and `Laziness' ($\mathcal{L}_{laz/S}$), such that,

\begin{equation}
    \small
    \mathcal{L}_{laz}(i,m,L(m))=\mathcal{L}_{laz/L}(i,L(m))+\mathcal{L}_{laz/S}(m).
\end{equation}

On one hand, $\mathcal{L}_{laz/L}$ forces Impatient Listener to guess the right candidate as soon as possible when reading the message $m$. For this purpose, with $i$ the ground-truth input and $L(m)=(L(m_{:0}),...,L(m_{:t}))$ the sequence of intermediate distributions, the Impatience Loss is defined as the mean cross-entropy loss between $i$ and the intermediate distributions:
\begin{equation}
    \mathcal{L}_{laz/L}(i,L(m))=\frac{1}{t+1}\sum_{k=0}^{t}\mathcal{L}_{std}(i,L(m_{:k})),
\end{equation}
Hence, all the intermediate distributions contribute to the loss function according to the following principle: the earlier the Listener predicts the correct output, the larger the reward is.

On the other hand, $\mathcal{L}_{laz/S}$ consists of an adaptive penalty on message lengths. 
The idea is to first let the system explore long and discriminating messages (\textbf{exploration step}) and then, once it reaches good enough communication performances, we apply a length cost (\textbf{reduction step}). With $|m|$ the length of the message associated with the input $i$ and `$\text{acc}$' the estimation of the accuracy (proportion of inputs correctly communicated weighted by appearance frequency), the Laziness Loss is defined as:

\begin{align}
    \mathcal{L}_{laz/S}(m)= \alpha(\text{acc})|m|
\end{align}

\begin{figure}[h]
        \centering
        \includegraphics[scale=0.75]{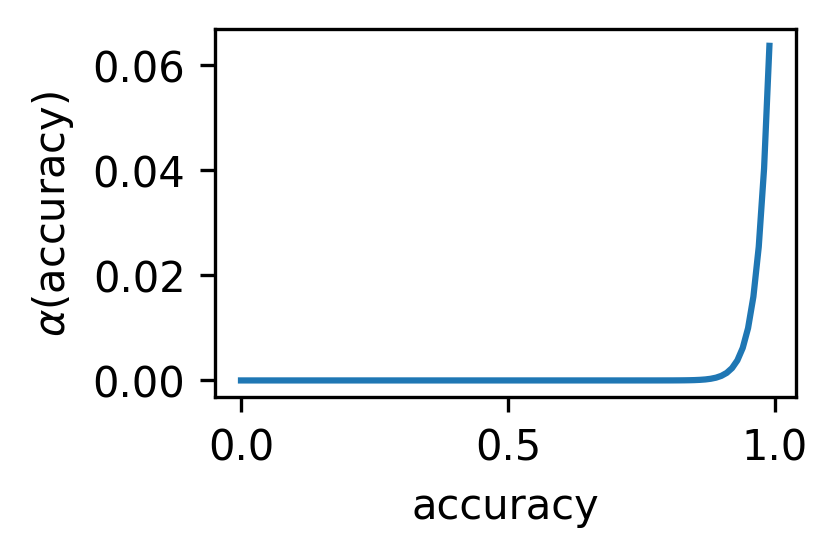}
        \caption{Scheduling of the regularization parameter $\alpha$ as a function of the accuracy. We distinguish two different regimes: the exploration and the reduction regimes. See the mathematical description in Appendix \ref{appendix:reg} }
        \label{fig:alpha_modelN}
\end{figure}

To schedule this two-step training, we model $\alpha$ as shown in Figure \ref{fig:alpha_modelN}. The regularization is mainly composed of two branches: (1) exploration step and (2) reduction step. The latter starts only when the two agents become successful.

\subsection{Optimization}
\label{section:framework:optim2}

The overall setting, which can be seen as a discrete auto-encoder, cannot be differentiated directly, as the latent space is discrete. 
We use a hybrid optimization between REINFORCE for Speaker \cite{williams:1992} and classic back-propagation for Listener \cite{schulman:2015}.

With $\mathcal{L}$ the loss of the system, $i$ the ground-truth input and $L(m)$ the output distribution of Listener that takes the message $m$ as input, the training task consists in minimizing the expectation of the loss $\mathbb{E}[\mathcal{L}(i,L(m))]$. The expectation is computed w.r.t the joint distribution of the inputs and the message sequences.  Let's denote $\theta_{L}$ and $\theta_{S}$ Listener and Speaker parameters respectively. The optimization task requires to compute the gradient $\nabla_{\theta_{S}\cup\theta_{L}}\mathbb{E}[\mathcal{L}(i,L(m))]$. An unbiased estimate of this gradient is the gradient of the following function:
\vspace{-0.1cm}
\begin{equation}
\label{eq:loss}
\small
 \mathbb{E}[\underbrace{\mathcal{L}(i,L(m;\theta_{L}))}_{(A)} + \underbrace{(\{\mathcal{L}(i,L(m;\theta_{L}))\} - b)\log P_{S}(m|\theta_{S})}_{(B)}], 
\end{equation}

where $\{.\}$ is the stop-gradient operation, $P_{S}(m|\theta_{S})$ the probability that Speaker generates the message $m$, $b$ the running-mean baseline used to reduce variance \cite{williams:1992}. We also promote exploration by encouraging Speaker's entropy \cite{williams:1991}. 

The gradient of (\ref{eq:loss}) w.r.t $\theta_{L}$ is found via conventional back-propagation (A) while gradient w.r.t $\theta_{S}$ is found with a REINFORCE-like procedure estimating the gradient via a Monte-Carlo integration calculated over samples of the messages (B). Once the gradient is estimated, it is eventually passed to the Adam optimizer \cite{kingma2014adam}. 

\bigbreak
In Appendix \ref{appendix:convergence}, we show that LazImpa leads to a stable convergence. We use the EGG toolkit \citep{Kharitonov:etal:2019} as a starting framework. For reproducibility, the code can be found at \url{https://github.com/MathieuRita/Lazimpa} and the set of hyper-parameters used is presented in Appendix \ref{appendix:experimental_settings}.

\section{Analytical method}
\label{section:method2}

As ZLA is defined informally, we first introduce reference distributions for comparison. Then, we propose some simple metrics to evaluate the overall efficiency of our emergent codes. Eventually, we provide a simple protocol to analyze the distribution of information within the messages.

\subsection{Reference distributions}
\label{section:method:ref2}

We compare the emergent languages to the reference distributions introduced in \citet{chaabouni:etal:2019}. We provide below a brief description of the different distributions, however, we invite readers to refer to the reference paper for more details.

\paragraph{Optimal Coding} \citep{Cover:Thomas:2006} guarantees the shortest average message length with \verb+max_len+  $=30$ and \verb+voc_size+ $=40$. To do so, we deterministically associate the shortest  messages to the most frequent inputs. See \citet{FerrerICancho:etal:2013} for more details about the derivation of Optimal Coding.

\paragraph{Natural Language} We also compare  emergent languages with several human languages. In particular, we consider the same languages of the reference paper (English, Arabic, Russian, and Spanish). These references consist of the mapping  from the frequency of the top $1000$ most frequent words  in each language to their length (approximated by the number of characters of each word).\footnote{We use the frequency lists from \url{http://corpus.leeds.ac.uk/serge/}.
}

\subsection{Efficiency metrics}
\label{section:method:eff_metrics2}

In this work, we examine the constraints needed for neural agents to develop efficient languages. We use three metrics to evaluate how efficient the different codes are.\\
For all metrics, $N$ denotes the total number of messages ($=$1000) and $l(m)$ the length of a message $m$.

\paragraph{Mean message length $L_{type}$:} measures the mean length of the messages assuming a \emph{uniform} weight for each input/message:
\begin{equation}
    L_{type}=\frac{1}{N}\sum_{m \in \mathcal{M}}l(m),
\end{equation}

\paragraph{Mean weighted message length $L_{token}$}: measures the average length of the messages weighted by their generation frequency:
\begin{equation}
    L_{token}=\sum_{m \in \mathcal{M}}p(m)l(m),
\end{equation}
where $p(m)$ is the probability of message $m$ (equal to the probability of input $i$ denoted by $m$) such that $\sum_{m \in \mathcal{M}}{p(m)}=1$. Formally, the message $m$ referring to the $k^{th}$ most frequent input would have a probability $\frac{1/k}{\sum_{1}^{1000}1/j}$. \\
Note that, the Optimal Coding is the one that minimizes $L_{token}$ \cite{Cover:Thomas:2006, FerrerICancho:etal:2013}.

\paragraph{ZLA significance score $p_{ZLA}$:} Let's note $(l_{i})_{i \in \mathcal{I}}$ a distribution of message lengths of a code. As a ZLA distribution is the one that minimizes $L_{token}$, we can check if $(l_{i})_{i \in \mathcal{I}}$ follows ZLA by testing if its $L_{token}$ is lower than any random permutation of its frequency-length mapping. This is the idea of the randomization test proposed by \citet{FerrerICancho:etal:2013}.

The test checks whether $L_{token}$ coincides with $\sum_{i \in \mathcal{I}} l_{i}f_{\sigma(i)}$, with $\sigma(i)$ a random permutation of inputs. We can eventually compute a p-value $p_{ZLA}$ (at threshold $\alpha$) that measures to which extent $L_{token}$ is likely to be smaller than any other weighted mean message length of a frequency-length mapping. $p_{ZLA}<\alpha$ indicates that any random permutation would have most likely longer weighted mean length. Thus $(l_{i})_{i \in \mathcal{I}}$ follows \emph{significantly} a ZLA distribution. 
Additional details are provided in Appendix \ref{appendix:quantif}.

\subsection{Information analysis}
\label{section:method:inf_metrics2}

We also provide an analytical protocol to evaluate how information is distributed within the messages. We consider a symbol to be informative if replacing it randomly has an effect on Listener's prediction. Formally, let's take the message $m=(m_{0},...,m_{t})$ associated to the ground truth input $i$ after training. To evaluate the information contained in the symbol at position $k$, $m_{k}$, we substitute it randomly by drawing another symbol $r_{k}$ uniformly from the vocabulary (except the \verb+EOS+ token). Then, we feed this new message $\Tilde{m}=(m_{1},...,r_{k},...,m_{t})$ into Listener that outputs $\Tilde{o}_{m,k}$ (index $m$ indicates that the original message was $m$, index $k$ indicates that the $k^{th}$ symbol of the original message has been replaced). We define $\Lambda_{m,k}$ a boolean score that evaluates whether the symbol replaced at position $k$ has an impact on the prediction, such that  $\Lambda_{k,m}=\mathbf{1}(\Tilde{o}_{m,k} \neq i)$. If $\Lambda_{m,k}=1$, the $k^{th}$ symbol of message $m$ is considered as informative. If $\Lambda_{m,k}=0$, it is considered as non-informative. We do not consider misreconstructed inputs, neither the position $t$, as $m_{t}$=\verb+EOS+.\footnote{As we only consider successful runs, more than 97\% of inputs are, by definition, well-reconstructed.} This token is needed for Listener's prediction at test time.

This test allows us to introduce some variables that quantify to which extent information is effectively distributed within the messages. As previously, we note $l(m)$ the length of message $m$ and $N$ the total number of messages.

\paragraph{Positional encoding} $(\Lambda_{.,k})_{1 \leq k \leq \verb+max_len+}$ : analyzes the position of informative symbols within an emergent code. We assign a score $\Lambda_{.,k}$ for each position $k$ that counts the proportion of informative symbols over all the messages of a language: 
    \begin{equation}
        \Lambda_{.,k}=\frac{1}{N(k)}\sum_{m\in \mathcal{M}}\Lambda_{m,k},
    \end{equation}
where $N(k)$ is the number of messages that have a symbol (different from \verb+EOS+) at position $k$.

\paragraph{Effective length} $L_{eff}$: measures the mean number of informative symbols by message:
    \begin{equation}
      L_{eff}=\frac{1}{N}\sum_{m\in \mathcal{M}} \sum_{k=1}^{l(m)-1} \Lambda_{m,k}.
    \end{equation}

$L_{eff}$ counts the average number of symbols Listener relies on (removing all the uninformative symbols for which $\Lambda_{m,k}=0$). A message with only informative symbols would have $L_{eff}=L_{type}-1$.\footnote{We subtract 1 as we disregard $\texttt{EOS}$ in all messages.}

\paragraph{Information density} $\rho_{inf}$  : measures the fraction of informative symbols in a language:
    \begin{equation}
        \rho_{inf}=\frac{1}{N}\sum_{m\in \mathcal{M}}\frac{1}{l(m)-1}\sum_{k=1}^{l(m)-1}\Lambda_{m,k}.
    \end{equation}
We integrate over the first $l(m)-1$ positions as we disregard \verb+EOS+ that occurs in all messages.\footnote{By convention, for the case where $m$=$(\texttt{EOS})$, $\frac{0}{0}$=$1$.} $0 \leq \rho_{inf} \leq 1$. If $\rho_{inf}=1$,  messages are limited to the informative symbols (all used by Listener to decode the message). The lower $\rho_{inf}$ is, the more non-informative symbols are in the message.

\bigskip
As we do not have Listener when generating Optimal Coding, we compute these metrics for the latter reference by considering all symbols, but $\texttt{EOS}$, informative. 

\section{Results}
\label{section:results}

\begin{table*}[h]
    \centering
    \begin{adjustbox}{max width=\textwidth}
    \begin{tabular}{|c|c||c|c|c||c|c|}
    \hline
         Class & Code & $L_{type}$ &$L_{token}$& $p_{ZLA}$ & $L_{eff}$ & $\rho_{inf}$ \\
        \hline
         Emergent & Standard Agents & $29.6 \pm 0.4$ & $29.91 \pm 0.07$ & $>1-10^{-5}$ & $3.33 \pm 0.46$ & $0.11 \pm 0.02$  \\  
          & LazImpa & $5.49 \pm 0.67$ & $3.78 \pm 0.34$ & $<10^{-5}$* &  $2.67 \pm 0.07$ & $0.60 \pm 0.07$ \\  
         \hline
         \hline
 
         References & Mean natural languages & $5.46 \pm 0.61$ & $3.55 \pm 0.14$ & $<10^{-5}$* & / & / \\
         & Optimal Coding & $2.96$ & $2.29$ & $<10^{-5}$* & $1.96$ & $1.00$ \\ 
         \hline
          
          \hline 
    \end{tabular}
    \end{adjustbox}
    \caption{Efficiency and information analysis of emergent codes and reference distribution. For each metric, we report the mean value and the standard deviation when relevant (across seeds when experimenting with emergent languages and across the natural languages presented in Section \ref{section:method:ref2} for  Mean natural languages). $L_{type}$ is the mean message length, $L_{token}$ is the mean weighted message length, $p_{ZLA}$ the ZLA significance score, $L_{eff}$ the effective length and $\rho_{inf}$ the information density. `/' indicates that the metric cannot be computed. For $p_{ZLA}$, `*' indicates that the p-value is significant ($<0.001$). }
    \label{tab:metrics}
\end{table*}

In this section, we study the code of our new communicative system, LazImpa, and compare it to the Standard Agents baseline and the different reference distributions. We show that LazImpa leads to near-optimal and ZLA-compatible languages. Eventually, we demonstrate how both Impatience and Laziness are required to get human-level efficiency. 
All the quantitative results of the considered codes are gathered in Table \ref{tab:metrics}.

\subsection{LazImpa vs.~Standard Agents}
\label{section:results:std2}
We compare here LazImpa to the baseline system Standard Agents both in terms of the length efficiency and the allocation of information.

\paragraph{Length efficiency of the communication.}

\begin{figure}[h] 
    \centering
    \includegraphics[scale=0.6]{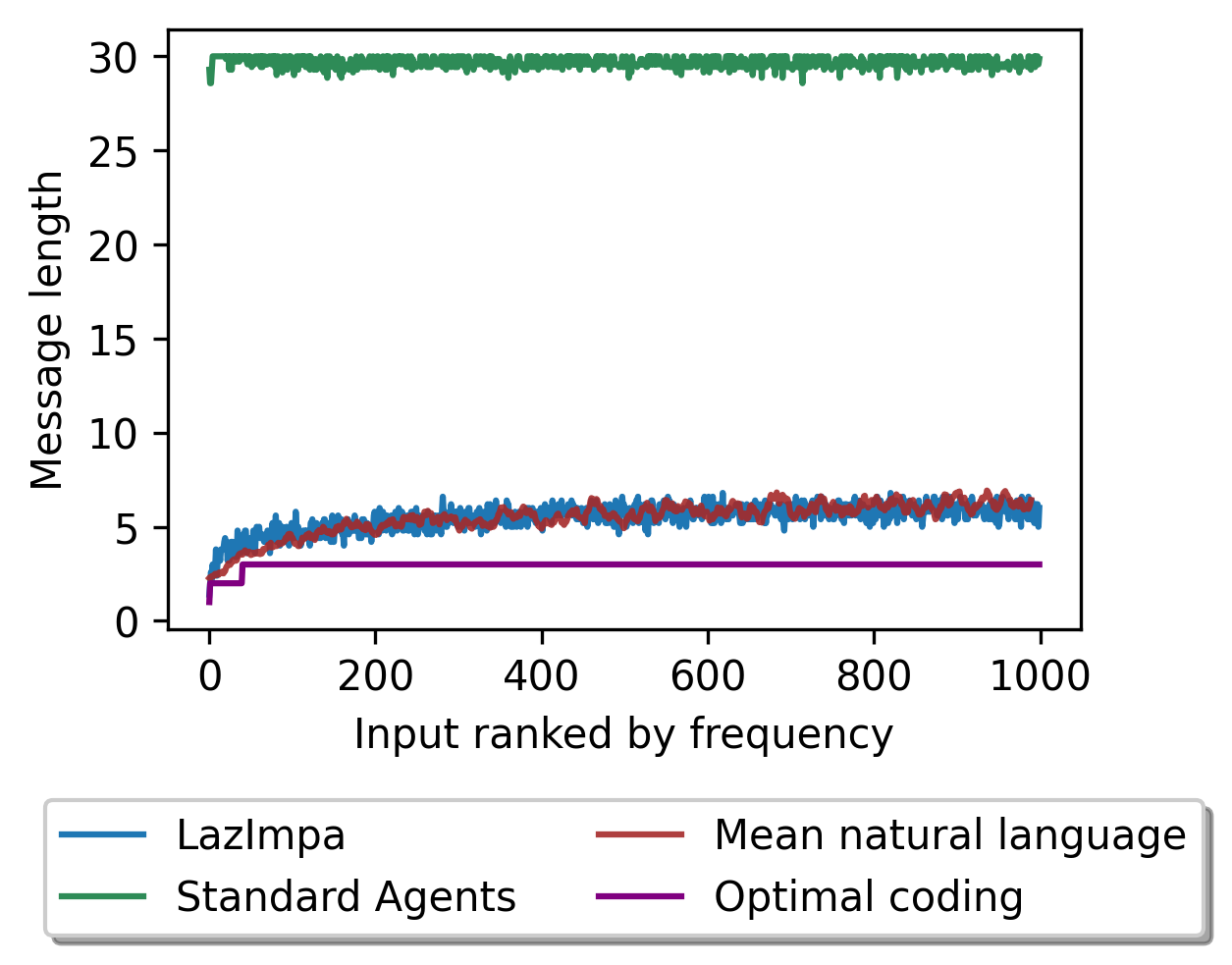}
    \caption{Average message length across successful runs as a function of input frequency rank.
    } 
    \label{fig:ZLAN}
\end{figure}

Contrary to Standard Agents, LazImpa develops an efficient communication as presented in Figure \ref{fig:ZLAN}. Indeed, its average length of the messages is significantly lower than the Standard Agents system (average $L_{type}$=$29.6$ for Standard Agents vs. $L_{type}$=$5.49$ for LazImpa). The latter demonstrates length distributions almost constant and close to the maximum length we set (=30). We demonstrate in Appendix \ref{appendix:learning_path} how the exploration  of long messages in Standard Agents is key for agents' success in the reconstruction game, even though, in theory, shorter messages are sufficient.  

Interestingly, both systems do not only differ by their average length, but also by the distribution of messages length. Specifically, the Standard Agents system follows significantly an anti-ZLA distribution (see Appendix \ref{appendix:quantif} for quantitative support of this claim) while LazImpa has an average $L_{token}$=$3.78$ showing a ZLA pattern: the shortest messages are associated to the most frequent inputs. The randomization test gives quantitative support of this observation ($p_{ZLA}<10^{-5}$).

\paragraph{Informativeness of the communication.}

\begin{figure*}[h]

    \begin{subfigure}[t]{.45\textwidth}
    \centering
    \includegraphics[width=0.8\textwidth]{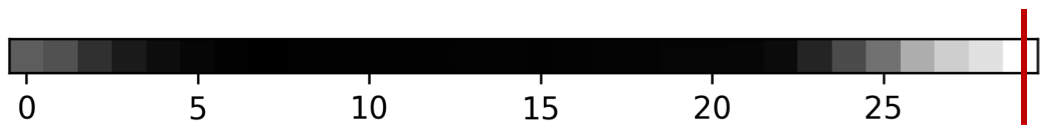}
    \caption{Standard Agents \vspace{.2cm}}
    \label{fig:posNN}
    \end{subfigure}
    \hspace{1.cm}
    \begin{subfigure}[t]{.45\textwidth}
    \centering
    \includegraphics[width=0.8\textwidth]{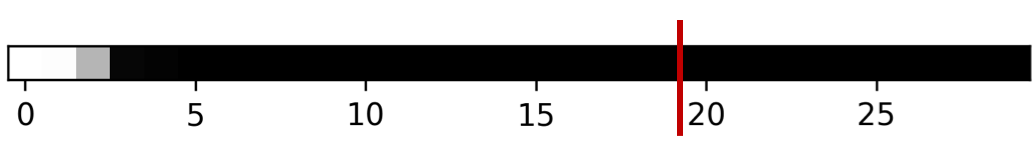}
    \caption{Standard Speaker + Impatient Listener}
    \label{fig:posILN}
    \end{subfigure}
     \\
    \hspace{1.cm}
    \begin{subfigure}[t]{.45\textwidth}
    \centering
    \includegraphics[width=0.8\textwidth]{figures/spectre_N_2.png}
    \caption{Lazy Speaker + Standard Listener}
    \label{fig:posNR}
    \end{subfigure}
    \hspace{1.cm}
    \begin{subfigure}[t]{.45\textwidth}
    \centering
    \includegraphics[width=0.8\textwidth]{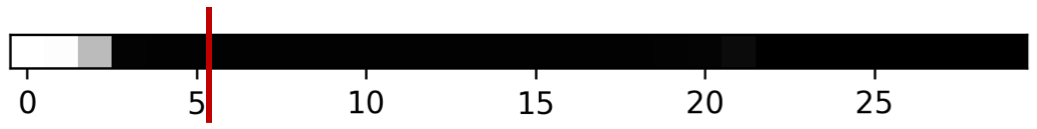}
    \caption{LazImpa \vspace{.2cm}}
    \label{fig:posIRLN}
    \end{subfigure}

\begin{subfigure}[t]{1\textwidth}
\centering
\vspace{-.1cm}
\includegraphics[width=0.18\textwidth]{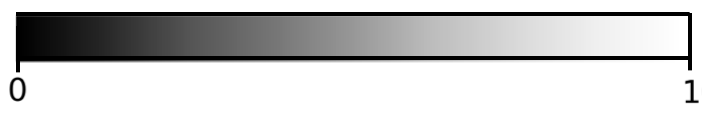}
\end{subfigure}
\caption{Fraction of informative symbols at each position k averaged across all emergent messages of successful runs ($(\Lambda_{k,.})_{0\leq k \leq29}$). Each box represents the proportion of informative symbols at a given position $\Lambda_{k,.}$ mapped to a color according to a gray gradient (black=0 ; white=1). The red vertical lines mark the mean message length $L_{type}$ across successful runs.}
    \label{fig:position_compaN}
\end{figure*}

When considering how Standard Agents system allocates information, shown in Figure \ref{fig:posNN}, we can make two striking observations. First,  only a very small part of the messages are informative (on average $\rho_{inf}=11\%$). Therefore, even if long messages seem necessary for the agents to succeed, most of the symbols are not used by Listener. In particular, if $L_{type}=29.6$ on average, the average number of symbols used by Standard Listener ($L_{eff}$) is only equal to $3.33$ (which is even smaller than natural languages' mean message length $L_{type}=5.46$). 
Surprisingly, we also observe that, if we restrict the messages to their informative symbols (i.e. removing positions $k$ with  $\Lambda_{k,.}=0$), the length statistics follow a ZLA-like distribution (see Figure \ref{fig:info_ZLA} in Appendix \ref{appendix:info_analysis}). Second, in all our experiments, the information is localized at the very end of the messages. That is, there is almost no actual information in the messages about Speaker's inputs before the last symbols. 

Contrarily, Figure~\ref{fig:posIRLN} shows a completely different spectrum for LazImpa. Indeed,  Impatient Listener relies on $\rho_{inf}=60\%$ of the symbols. This corresponds to a big increase compared to $\rho_{inf}=19\%$ when using Standard Agents. Yet, we are still far from the $100\%$ observed in Optimal Coding. That is, even with the introduction of a length cost (with Lazy Speaker), we still encounter non-informative symbols. Finally, these informative symbols are localized in the first positions, opposite to what we observed with Standard Agents. We will show in Section~\ref{section:results:compamodels2} how this immediate presence of information is crucial for  the length reduction of the messages.

\bigbreak
In sum, if we consider only \emph{informative/effective} positions, Standard Agents use efficient and  ZLA-like (effective) communicative protocol. However, they make it maximally long adding non-informative symbols at the beginning of each message. Introducing LazImpa reverses the length distribution. Indeed, we observe with LazImpa the emergence of efficient and ZLA-obeying languages, with significantly larger $\rho_{inf}$.

\subsection{LazImpa vs. reference distributions}
\label{section:results:rf2}

We demonstrated above how LazImpa leads to codes with length significantly shorter than the one obtained with Standard Agents. 

We compare it here with stricter references, namely natural languages and Optimal Coding. We show that LazImpa results in languages as efficient as natural languages both in terms of length statistics and symbols distribution. However, agents do not manage to reach optimality.
\paragraph{Comparison with natural languages. }

\begin{figure*}[h]
    \begin{subfigure}[t]{.46\textwidth}
    \centering
    \includegraphics[scale=0.55]{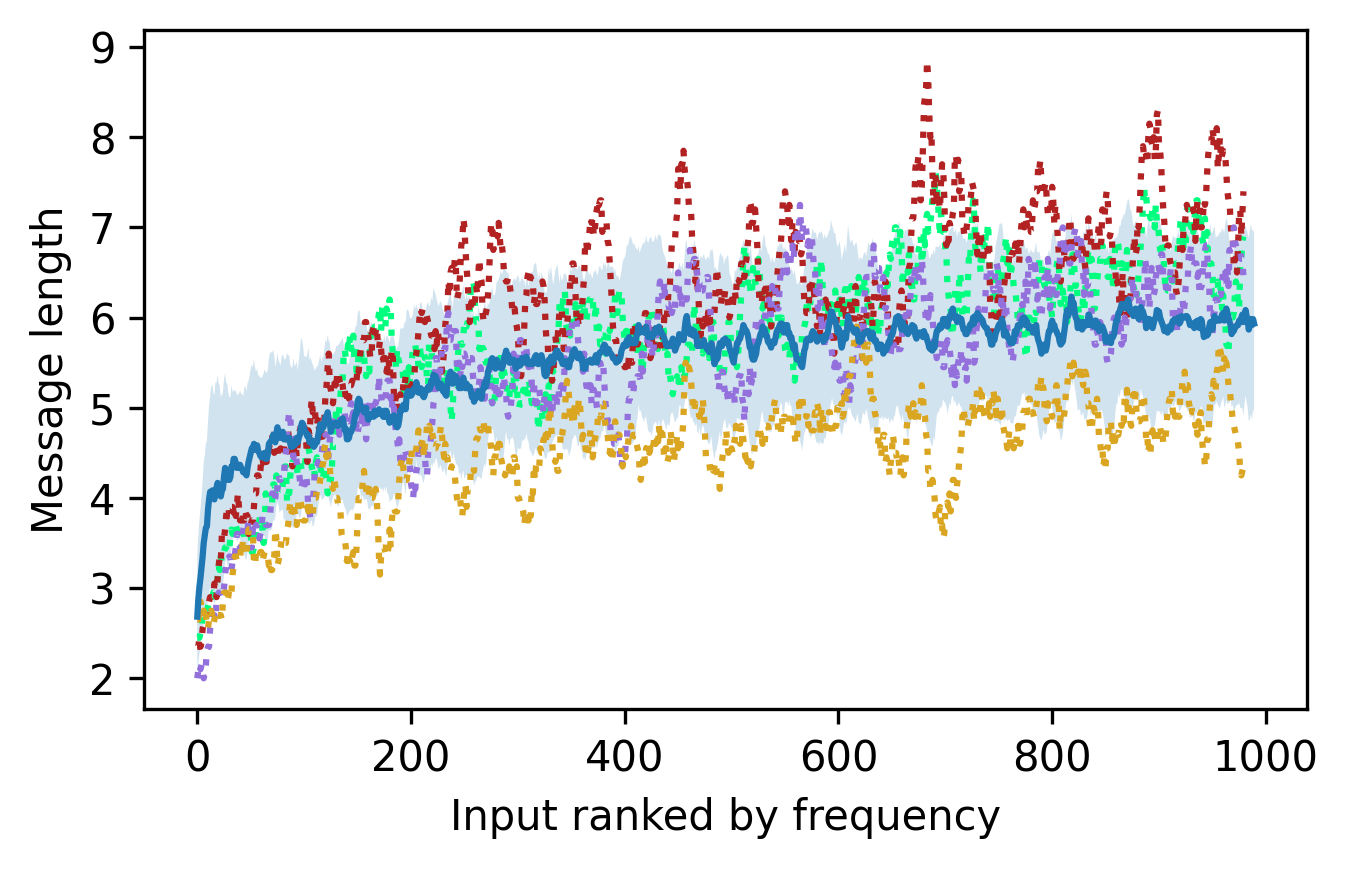}
    \end{subfigure}
    \hfill
    \begin{subfigure}[t]{.46\textwidth}
    \centering
    \includegraphics[scale=0.55]{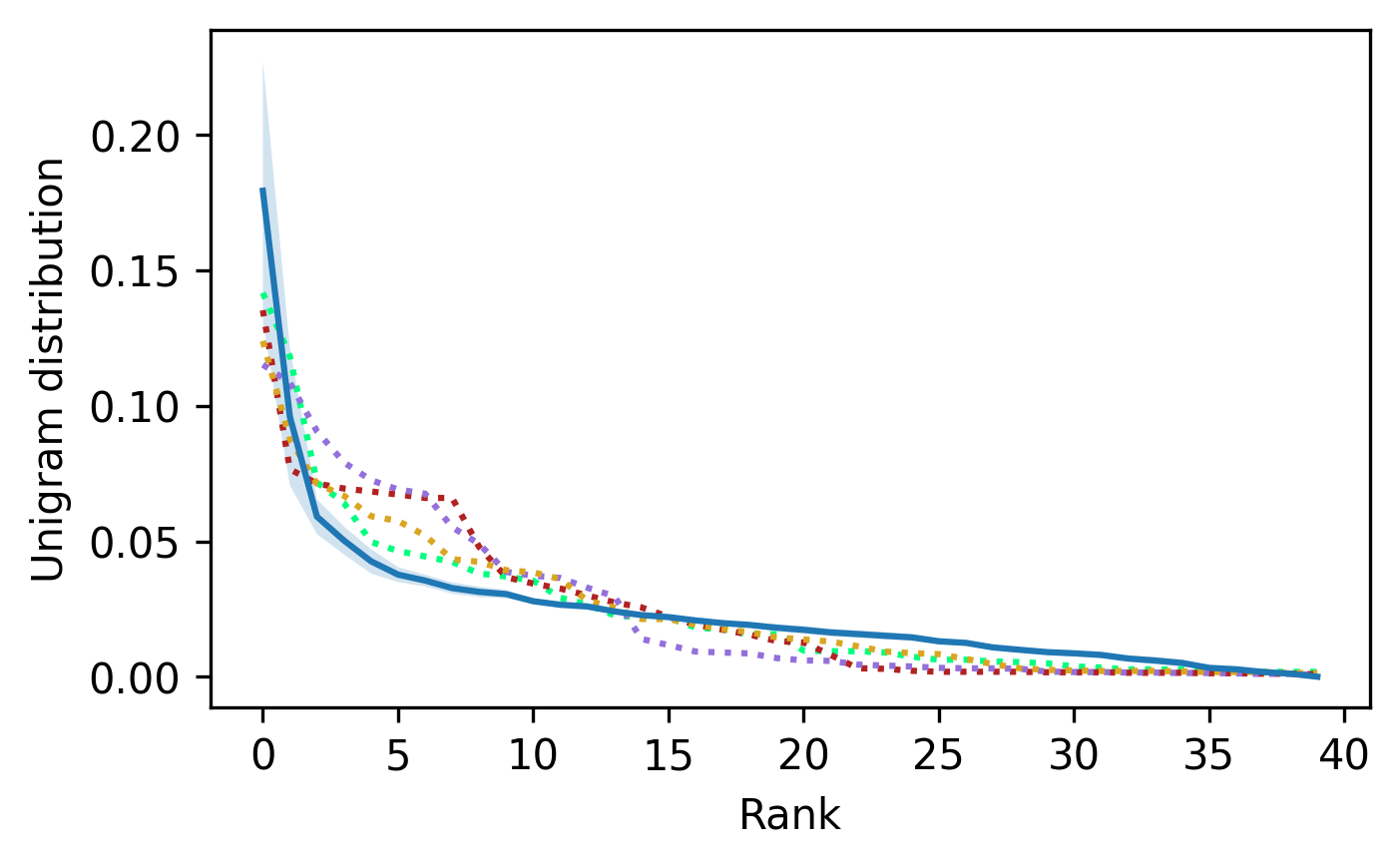}
    \end{subfigure}
    \begin{subfigure}{1.\textwidth}
    \centering
    \includegraphics[scale=0.7]{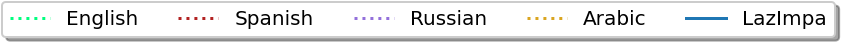}
    \end{subfigure}
    \\
    \begin{subfigure}{.46\textwidth}
    \caption{Message length of natural languages and LazImpa (averaged across successful runs) as a function of input frequency rank. For readability, the curves have been smoothed using a sliding average of 20 consecutive lengths, see the real curves in Appendix \ref{appendix:stat_NL}. The light blue interval shows 1 standard deviation for LazImpa's distribution.}
    \label{fig:compa_lenN}
    \end{subfigure}
    \hfill
    \begin{subfigure}{.46\textwidth}
    \caption{Unigrams distribution of natural languages and LazImpa (averaged across successful messages) ranked by unigram frequency. The light blue interval shows 1 standard deviation for LazImpa's unigrams distribution.}
    \label{fig:compa_uniN}
    \end{subfigure}
    \vspace{-.5cm}
    \caption{Comparison of LazImpa' statistics and natural languages.}
    \label{fig:compa_NLN}
\end{figure*}

We see in Figure \ref{fig:compa_lenN} that the message lengths in the emergent communication are analogous to the words lengths in natural languages: close average $L_{token}$ and $L_{type}$ (see Table~\ref{tab:metrics}). 

We further compare their unigram distributions. \citet{chaabouni:etal:2019} showed that Standard Agents develop repetitive messages with a skewed unigram distribution. Our results, in Figure \ref{fig:compa_uniN}, show that, on top of a ZLA-like code, LazImpa enables the emergence of natural-language-like unigram distribution, without any particular repetitive pattern. Intriguingly, this similarity with natural languages is an unexpected property as a uniform distribution of unigrams would lead to a more efficient protocol.

\paragraph{Comparison with Optimal Coding. }

If LazImpa leads to significantly more efficient languages compared to Standard Agents, these emergent languages are still not as efficient as Optimal Coding (see Figure \ref{fig:ZLAN}). One obvious source of sub-optimality is the addition of uninformative symbols at the end of the messages (i.e. the difference between $L_{eff}$=$2.67$ and $L_{type}$-$1$=$4.49$). Interestingly, when analyzing the intermediate predictions of Impatient Listener, we see that this model is actually able to guess the right input only reading approximately the $L_{eff}$ first positions (see Appendix \ref{appendix:min_length} for details). However, we still can note that the informative length $L_{eff}$ is slightly sub-optimal ($L_{eff}=2.67$ for LazImpa, $L_{eff}=1.96$ for Optimal Coding). This difference can be explained by the non-uniform use of unigrams. Specifically, we show in Appendix \ref{appendix:min_length} that effective lengths of LazImpa messages approximate Optimal Coding when the latter uses the same skewed unigram distribution.

\subsection{Ablation study}
\label{section:results:compamodels2}

\begin{figure*}[h]
    \begin{subfigure}[t]{.46\textwidth}
    \centering
    \includegraphics[scale=0.245]{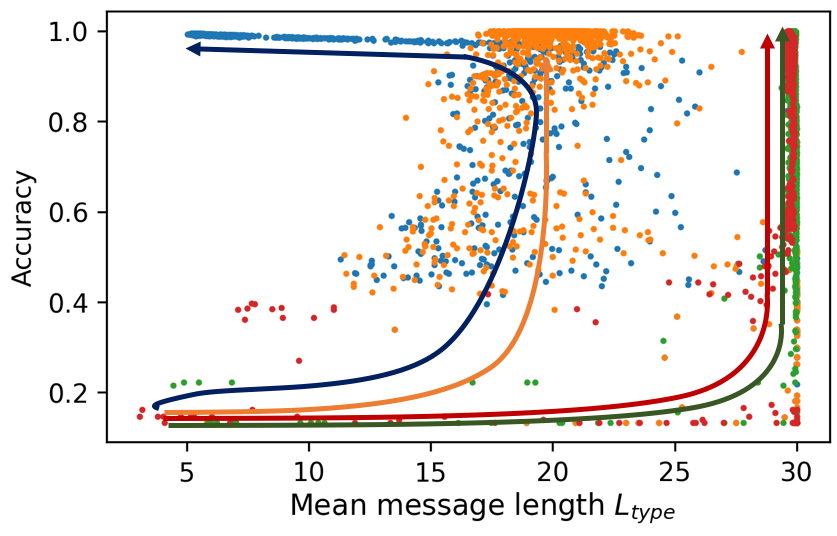}
    \end{subfigure}
    \hfill
    \begin{subfigure}[t]{.46\textwidth}
    \centering
    \includegraphics[scale=0.55]{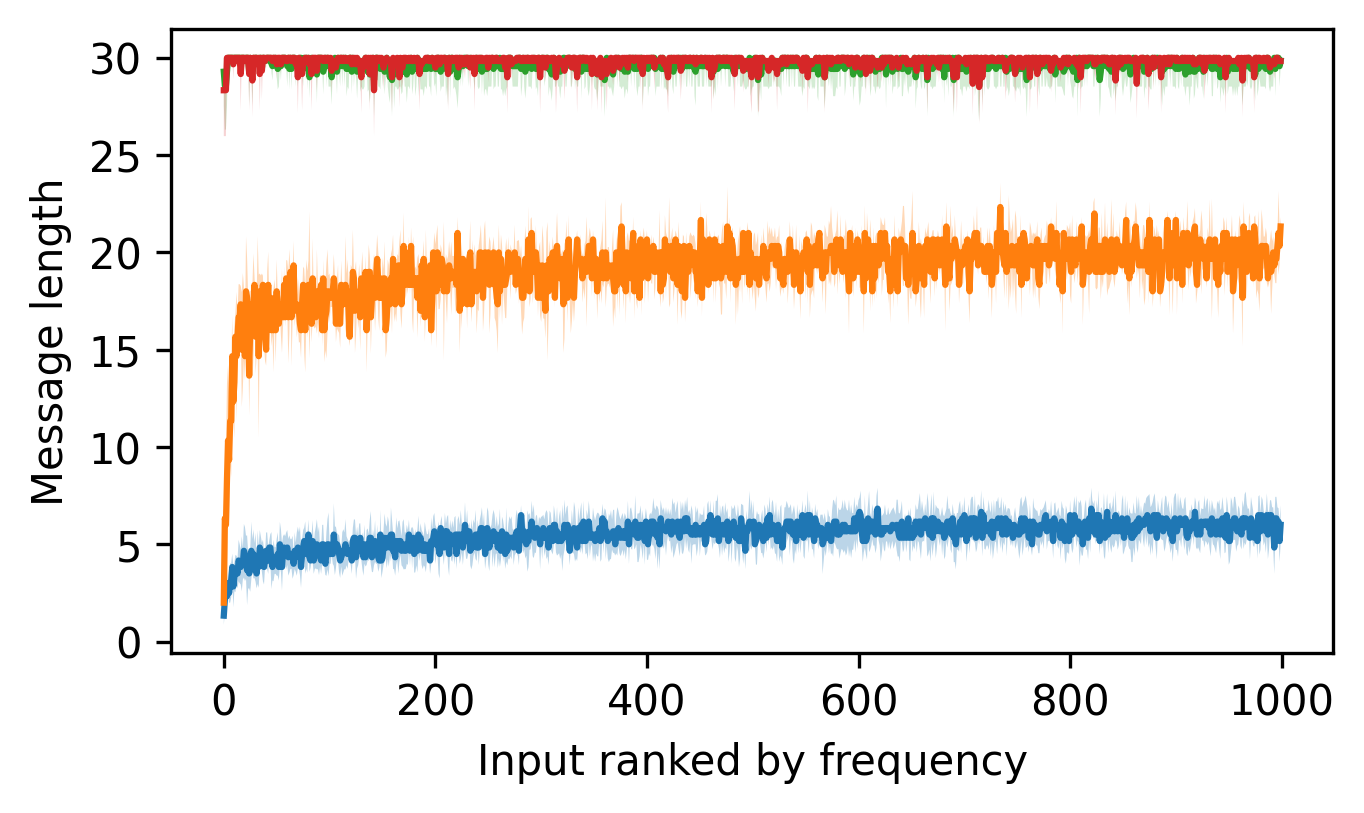}
    \end{subfigure}
    \begin{subfigure}{1.\textwidth}
    \centering
    \includegraphics[width=0.92\textwidth]{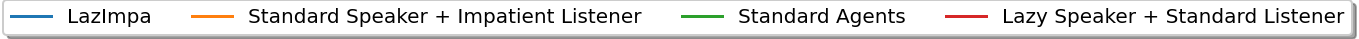}
    \end{subfigure}
    \\
    \begin{subfigure}{.46\textwidth}
    \caption{Joint evolution of the accuracy and mean length for the different models. Each point shows the couple ($L_{type}$,accuracy) of one training episode. Arrows represent the average joint evolution of the two variables.}
    \label{fig:compa_learning_pathN}
    \end{subfigure}
    \hfill
    \begin{subfigure}{.46\textwidth}
    \caption{Average message length as a function of input frequency rank for the different systems. Light color intervals show 1 standard deviation.}
    \label{fig:compa_lengthN}
    \end{subfigure}
    \vspace{-.5cm}
    \caption{Comparison of different communication systems.
    } 
\end{figure*}

We have just seen that our new communication system LazImpa allows agents to develop an efficient and ZLA-obeying language whose statistical properties are close to those of natural languages. In this section, we analyze the effects of the modeling choices we have made.

We first look at the effect of Laziness. To do so, we compare LazImpa to the system ``Standard Speaker + Impatient Listener'' (i.e. removing the length regularization). Figure \ref{fig:compa_learning_pathN} shows the joint evolution of the mean length of messages ($L_{type}$) and game accuracy. We observe that our non-regularized system, similarly to LazImpa, initially explores long messages  while being more successful (exploration step). Surprisingly, even in the absence of Laziness, the exploration step does not continue to maximally long messages, as it is the case for Standard Agents, but breaks at length $\approx 20$. However, \emph{contrary to LazImpa}, ``Standard Speaker + Impatient Listener'' does not present a reduction step (a reduction of mean length for a fixed good accuracy). Thus, as expected, the introduction of Laziness in LazImpa is responsible for the reduction step, and hence for a shorter and more efficient communication protocol. However, we note in Figure \ref{fig:compa_lengthN}, that Impatience alone is sufficient for the emergence of ZLA. Moreover, when looking at the information spectrum, comparing ``Standard Speaker + Impatient Listener'' (Figure \ref{fig:posILN}) to LazImpa (Figure \ref{fig:posIRLN}), we observe how alike both systems allocate information and differ only by their mean length.

Second, we investigate the role of Impatience. We see in Figure \ref{fig:compa_learning_pathN} that the system ``Lazy Speaker + Standard Listener'' admits a visually different dynamic compared to LazImpa. In particular, the exploration step leads to significantly longer messages, close to \verb+max_len+. Interestingly, if we demonstrated above the necessity of Laziness for the reduction step, alone, it does not induce it: no reduction step in the ``Lazy Speaker + Standard Listener'' system is observed. This is due to the necessity of long messages when experimenting with Standard Listener. Specifically, as informative symbols are present only at the last positions (see Figure \ref{fig:posNR}), introducing a length regularization provokes a drop in accuracy, which in turn cancels the regularization. In other words, the length regularization scheduling stops at the exploration step, which makes the system almost equivalent to Standard Agents (this could be also seen experimentally  in Figures \ref{fig:compa_learning_pathN} and \ref{fig:compa_lengthN}).

Taken together, our analysis emphasizes the importance of both Impatience \emph{and} Laziness for the emergence of efficient communication.

\section{Conclusion}

We demonstrated that a standard communication system, where standard Speaker and Listener LSTMs are trained to solve a simple reconstruction game, leads to long messages, close to the maximal threshold. Surprisingly, if these messages are long, LSTM agents rely only on a small number of informative message symbols, located at the end. We then introduce LazImpa, a constrained system that consists of \emph{Laz}y Speaker and \emph{Impa}tient Listener. On the one hand, Lazy Speaker is obtained by introducing a cost on messages length once the communication is successful. We found that early exploration of potentially long messages is crucial for successful convergence (similar to the exploration in RL settings). On the other hand, Impatient Listener aims to succeed at the game as soon as possible, by predicting Speaker's input at each message's symbol.

We show that both constraints are \emph{necessary} for the emergence of a ZLA-like protocol, as efficient as natural languages. Specifically, Lazy Speaker alone would fail to shorten the messages. We connect this to the importance of the Impatience mechanism to locate useful information at the beginning of the messages.
If the function of this mechanism is subject to a standing debate \cite[e.g.,][]{jackendoff2007,anderson2013}, many prior works had pointed to its necessity to human language understanding \citep[e.g.,][]{friston:2010,clark:2013}. We augment this line of works and suggest that impatience could be at play in the emergence of ZLA-obeying languages. However, if impatience leads to ZLA, it is not sufficient for human-level efficiency. In other words, efficiency needs constraints \emph{both} on Speaker and Listener sides. 

Our work highlights the importance of introducing the right pressures in the communication system. Indeed, to construct automated agents that would eventually interact with humans, we need to introduce task-agnostic constraints, allowing the emergence of more human-like communication. Moreover, while being general, LazImpa provides a more stable optimization compared to the unconstrained system. Finally, this study opens several lines of research. One would be to investigate further the gap from optimality. Indeed, while LazImpa emergent languages show human-level efficiency, they do not reach optimal coding. Specifically, emergent languages still have non-informative symbols at the end of the messages. If these additional non-useful symbols drift the protocol from optimality, we encounter similar trend in human \cite{marslen:1987} and animal communication  \cite{mclachlan2020speedy}. We leave the understanding of the role of these non-informative symbols and how we can reach optimal coding for future works. A second line of research would be to apply this system to other games or NLP problems and study how it affects other properties of the language such as regularity or compositionality.

\section*{Acknowledgments}
We would like to thank Emmanuel Chemla, Marco Baroni, Eugene Kharitonov, and the anonymous reviewers for helpful comments and suggestions.

This work was funded in part by the European Research Council (ERC-2011-AdG-295810 BOOTPHON), the Agence Nationale pour la Recherche (ANR-17-EURE-0017 Frontcog, ANR-10-IDEX-0001-02 PSL*, ANR-19-P3IA-0001 PRAIRIE 3IA Institute) and grants from CIFAR (Learning in Machines and Brains), Facebook AI Research (Research Grant), Google (Faculty Research Award), Microsoft Research (Azure Credits and Grant), and Amazon Web Service (AWS Research Credits).

\bibliography{bib}
\bibliographystyle{acl_natbib}

\newpage 
\bigbreak
\mathieu{a}
\bigbreak
\newpage

\appendix

\setcounter{figure}{6}
\setcounter{table}{1}
\setcounter{equation}{9}
\section{Appendix}

\subsection{Experimental settings}
\label{appendix:experimental_settings}

\subsubsection{Input space}
\label{appendix:input_space}

\begin{figure}[h]
    \centering
    \includegraphics[scale=0.6]{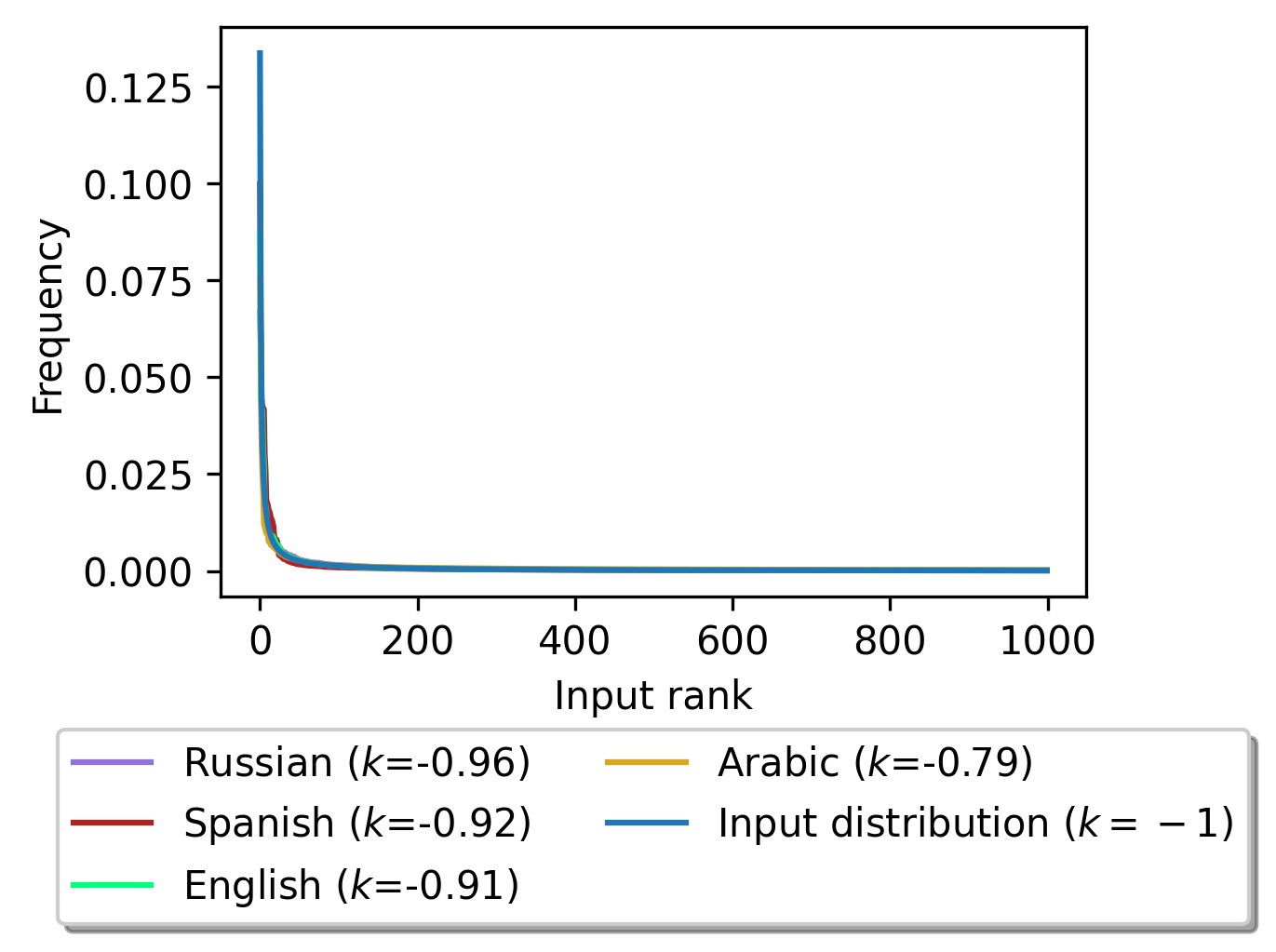}
    \caption{Comparison between the input distribution of our artificial environment and the distribution of the 1000 most frequent words  in different natural languages (the coefficient $k$ refers to the coefficient of the power-law for each language when fitted by a linear regression).}
    \label{fig:input_distrib}
\end{figure}

The input space $\mathcal{I}$ is composed of 1000 one-hot vectors. Each of them has to be communicated by Speaker to Listener. In order to fit the distribution of words in natural languages, the inputs are fed from a power-law distribution. Indeed, as demonstrated in Figure \ref{fig:input_distrib}, distribution of words in natural languages follow power-laws with exponents $k$ between $-0.79$ (Arabic) and $-0.96$ (Russian). In our experiment, we choose $k=-1$. 

\subsubsection{Agents}

In all our experiments, we fix the architecture of the agents. Speaker is a 1-layer LSTM \cite{hochreiter1997} with a hidden size equal to 100. Listener is also a 1-layer LSTM with a hidden size equal to 600.

\subsubsection{Optimization}

For the training, we use the Adam optimizer \cite{kingma2014adam} with a learning rate equal to 0.001. We train the agents for 1500 epochs. During one episode, the system is fed with 100 batches of 512 inputs sampled with replacement from the power-law distribution. In addition, we enforce exploration with an entropy regularization coefficient equal to 2 \cite{williams:1991}.

To ensure the robustness of our results, we ran the experiments with 6 different random seeds. All the experiments have been successful, i.e.~they reach an accuracy of 99\%. This accuracy is weighted by the frequency of inputs. On average, more than 97.5\% of inputs are well communicated.

\subsubsection{Adaptive regularization coefficient}
\label{appendix:reg}

As defined in the main paper, the adaptive regularization coefficient is scheduled as a function of the accuracy in order to have the following two-step scheme:
\begin{itemize}
    \item \textbf{Exploration step}: during the first part of the training (low accuracy), the regularization coefficient is almost null
    \item \textbf{Reduction step}: Once the communication becomes successful (high accuracy), we start introducing a regularization.
\end{itemize}

A fair equation to model this two-step scheme is:

\begin{equation}
    \alpha(\text{accuracy})=\frac{\text{accuracy}^{\beta_{1}}}{\beta_{2}}
\end{equation}

where $(\beta_{1},\beta_{2}) \in \mathbb{R}^{2}$ is a new couple of hyper-parameters. Intuitively, the two parameters allow to control (a) the threshold from which the regularization becomes effective (with $\beta_{1}$) and (b) the intensity of the regularization (with $\beta_{2}$). In our experiments, we introduce a late regularization choosing: $\beta_{1}=45$. We set $\beta_{2}=10$ in order to enables the system to reach an accuracy close to 1.
\\
Note that other regularization scheduling can be applied. The only requirement is that the agents successfully communicate before the start of the reduction step.

\subsection{Characterization of the emergent communication with Standard Agents}
\label{appendix:characterization_emergent}

In this section, we report complements about the characterization of the emergent communication with Standard Agents.

\subsubsection{Quick use of long messages}
\label{appendix:learning_path}

\begin{figure}[!h]
    \centering
    \includegraphics[scale=0.6]{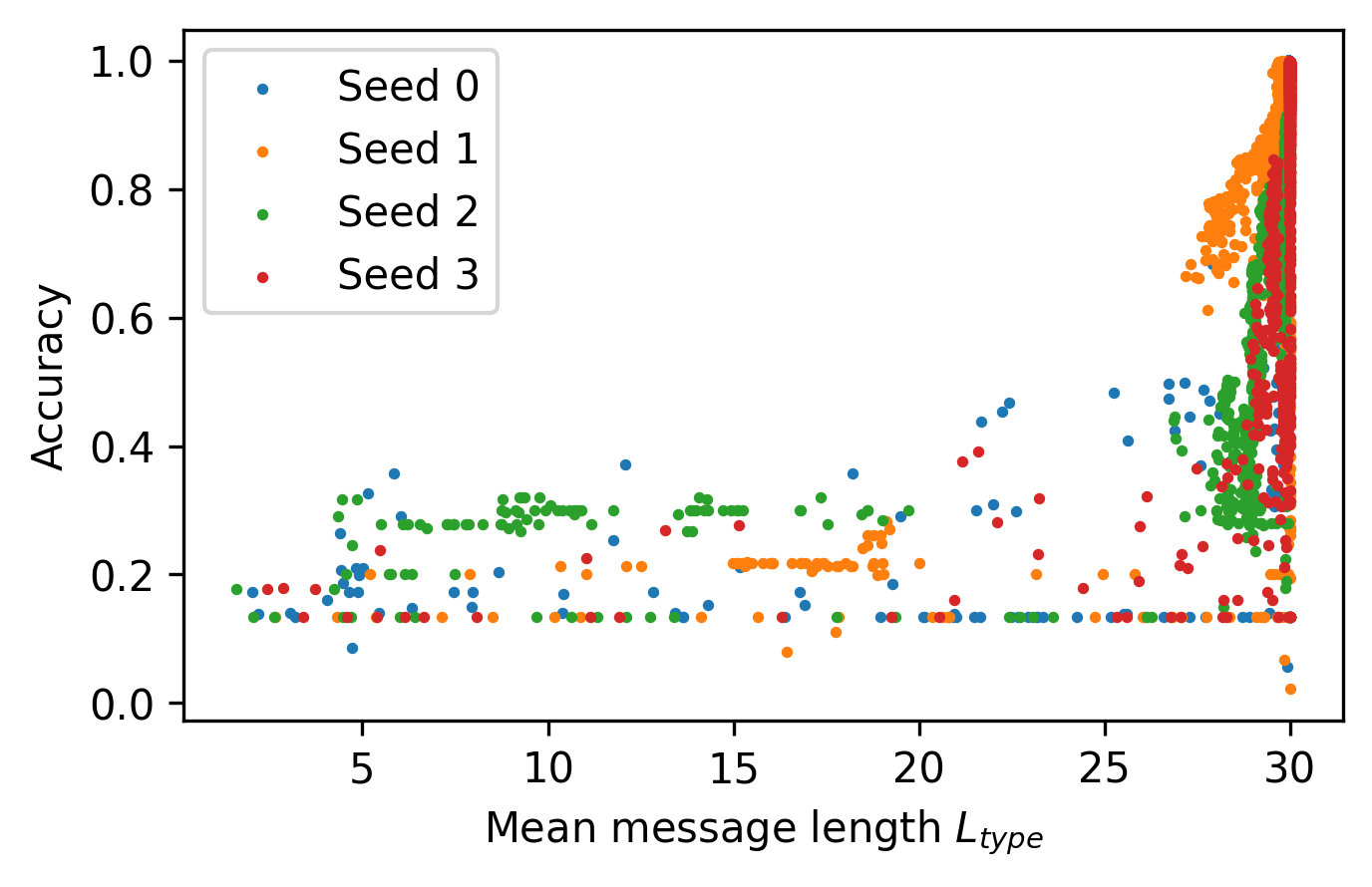}
    \caption{Accuracy as a function of the mean length for 4 different seeds. Each point represents a couple $(\text{accuracy},\text{mean length})$.}
    \label{fig:normal_learning_path}
\end{figure}

To bring more insights about the length inefficiency observed in the main paper, we characterize each episode by the couple accuracy (i.e.~the  proportion of inputs correctly communicated by the agents weighted by the frequency of appearance) and mean length (i.e.~the average length of the messages generated by the Speaker).

During the training time, we analyze how this couple evolves. The results with four randomly selected seeds are shown in Figure \ref{fig:normal_learning_path}. As we can see, at the beginning of the learning process (low accuracies), both the mean length of the messages and the accuracy are quite low (the lowest accuracy value 0.13 corresponds to the good prediction of the most frequent input). Then, the mean message length is increasing without a strong effect on the accuracy. It is only when the agents start to use long messages (higher than 25 for a maximum length of 30) that the communication becomes successful.
Therefore, we see that exploration of long messages seems key for the agents to reach high accuracies. 

\subsubsection{Efficient \emph{informative} symbols}
\label{appendix:info_analysis}

\begin{figure}[h]
    \centering
    \includegraphics[scale=0.65]{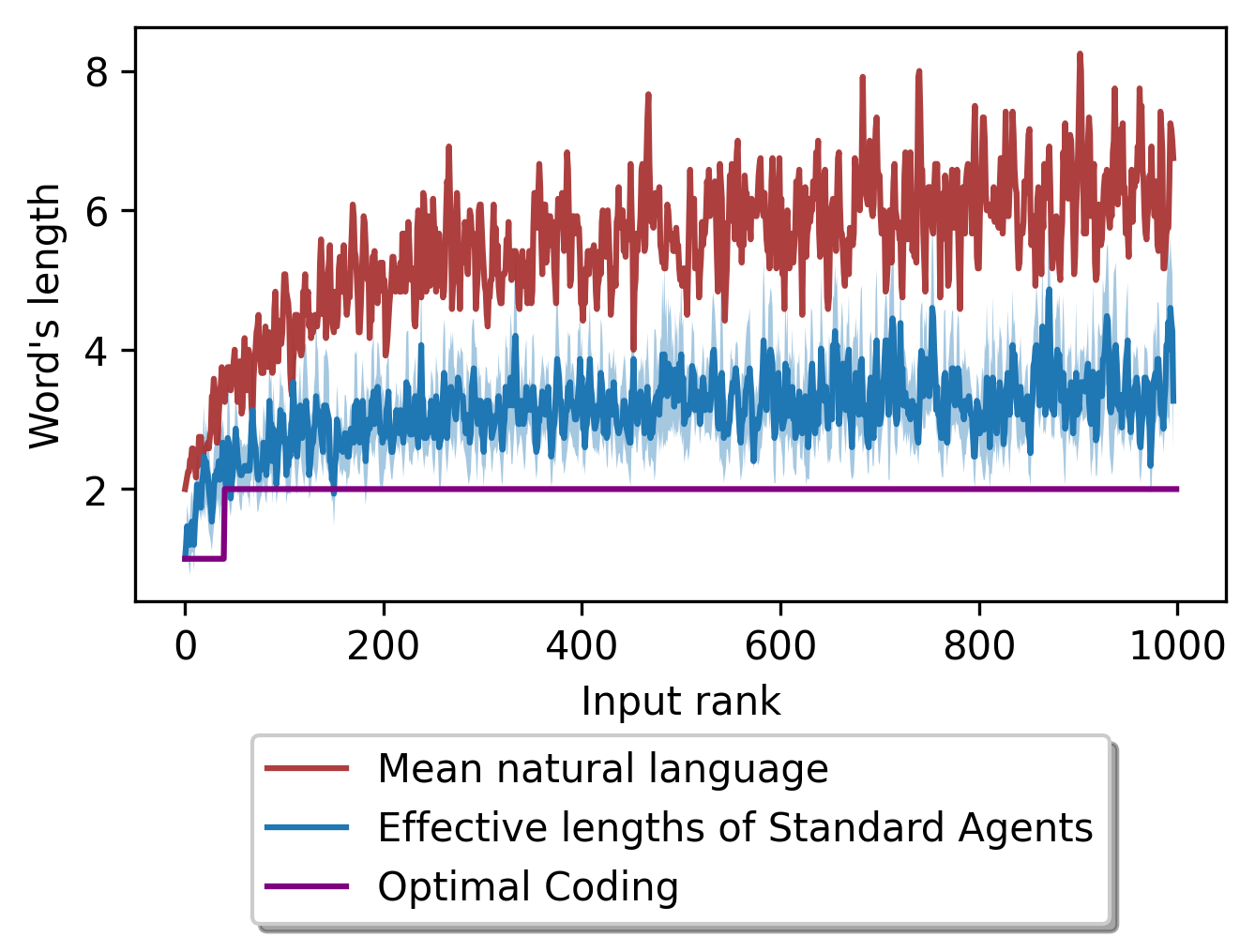}
    \caption{ Average length distribution of informative parts in Standard Agents code compared to the mean words distribution of natural languages and Optimal Coding. The light blue interval shows 1 standard deviation. For readability, the natural language distribution have been smoothed with a sliding average of 3 consecutive lengths.}
    \label{fig:info_ZLA}
\end{figure}

We analyze the statistical properties of the informative parts of the messages that emerge from Standard Agents. As defined in the main paper, we consider a symbol informative if it is used by Listener for the reconstruction. We remove all the non-informative symbols from the messages (i.e. positions $k$ with $\Lambda_{k,.}=0$). In Figure \ref{fig:info_ZLA}, we plot the length of informative parts of messages associated to inputs ranked by frequency (average distribution over the different runs). We compare it to the average words length distribution of natural languages and to Optimal Coding. As we can see in the figure, even though Standard Agents produce an inefficient code (as seen in the main paper) the length statistic of the informative parts is close to Optimal Coding. Interestingly, we even note an emergent code more efficient than natural languages. 
In addition, even if no constraint is applied on informative parts, we observe that it follows ZLA.

\subsection{Comparing communication systems}
\label{appendix:model_compa}

\subsubsection{Convergence}
\label{appendix:convergence}

\begin{figure*}[h]
    \centering
    \begin{subfigure}[t]{.3\textwidth}
    \includegraphics[scale=0.5]{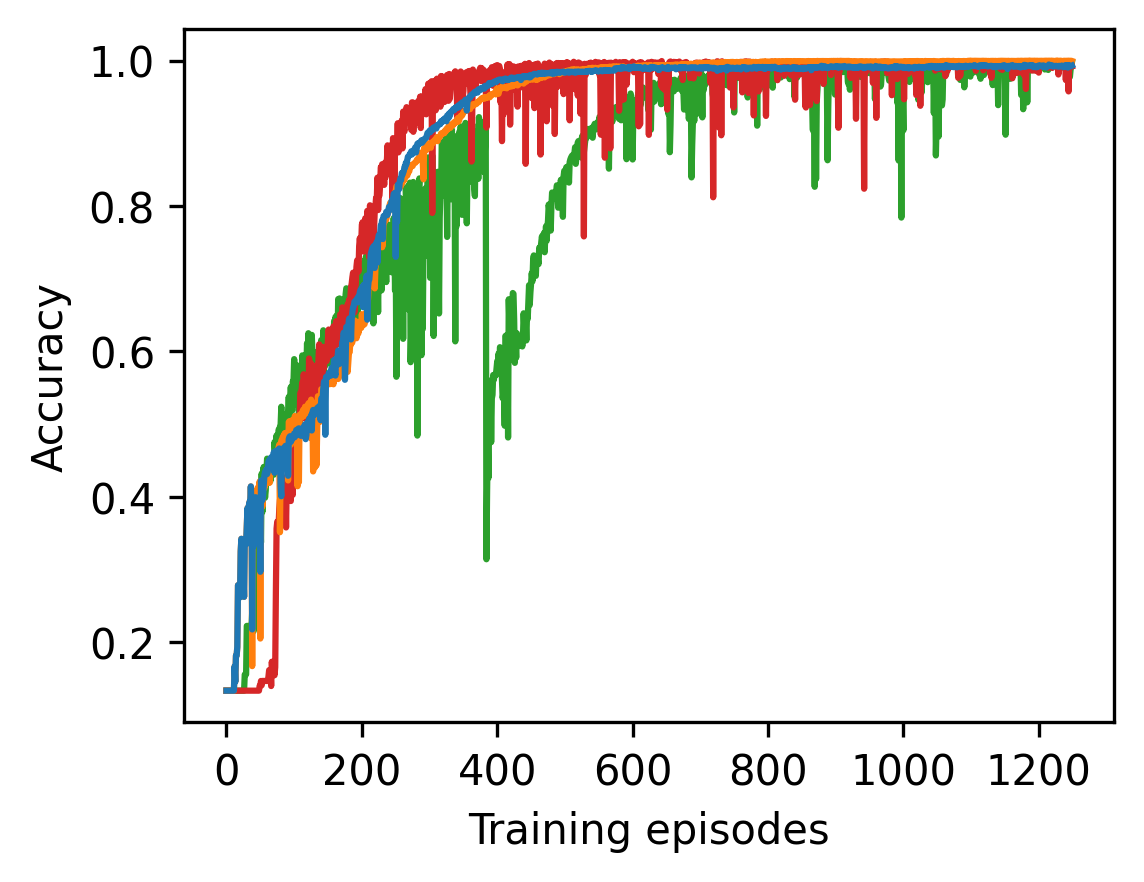}
    \caption{Seed 1}
    \end{subfigure}
    \begin{subfigure}[t]{.3\textwidth}
    \includegraphics[scale=0.5]{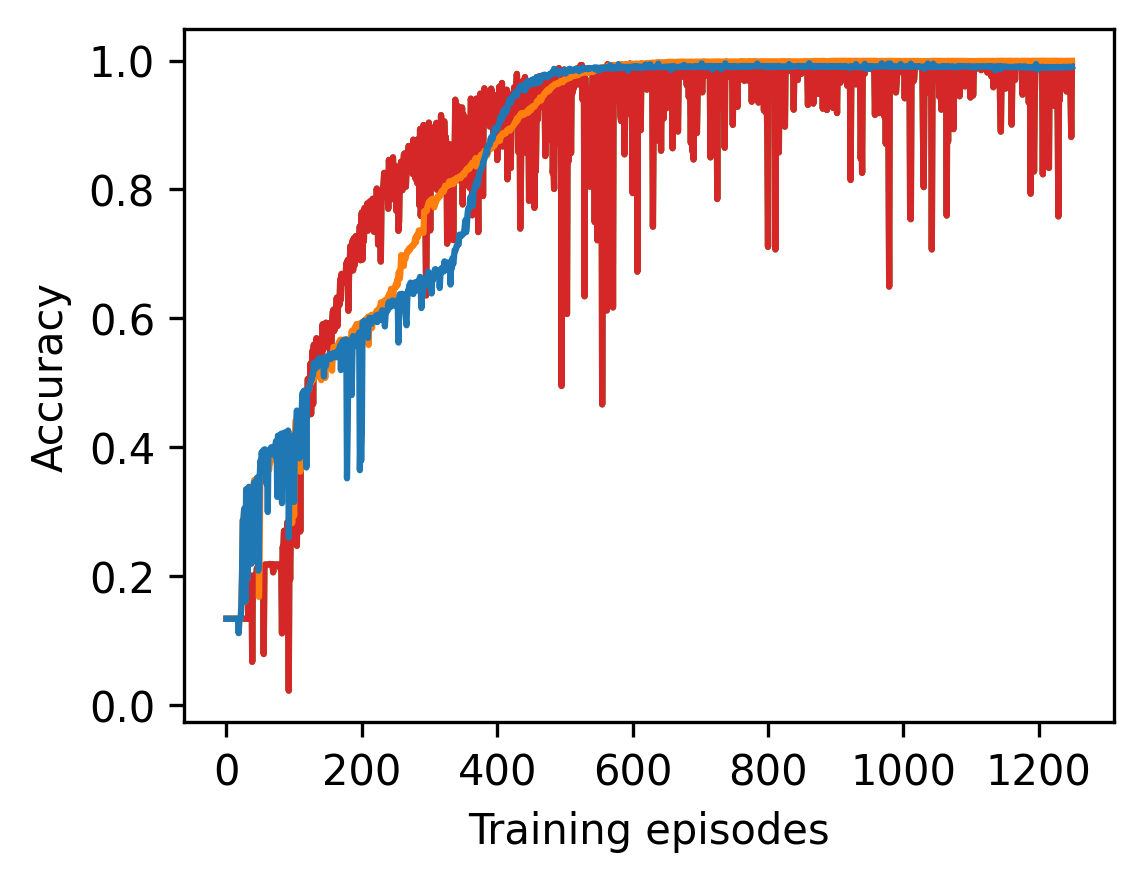}
    \caption{Seed 2}
    \end{subfigure}
    \begin{subfigure}[t]{.3\textwidth}
    \includegraphics[scale=0.5]{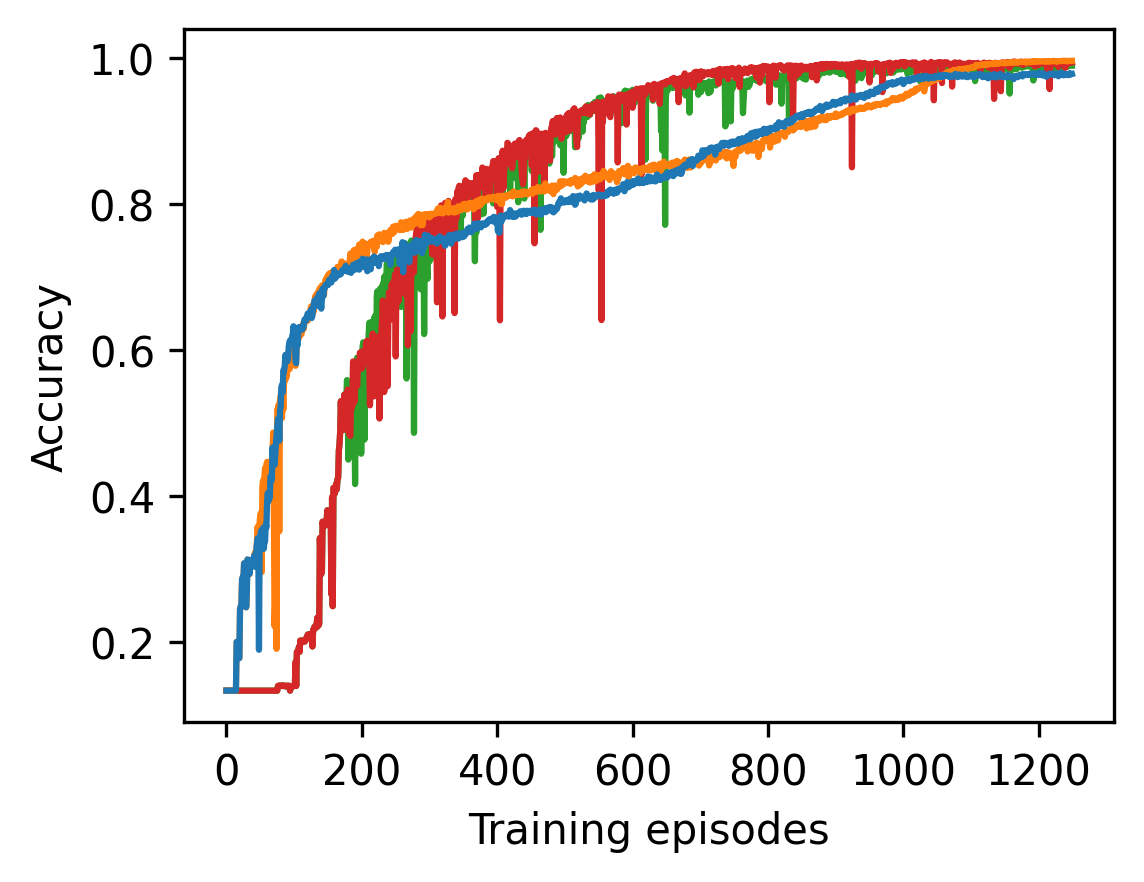}
    \caption{Seed 3}
    \end{subfigure}
    \begin{subfigure}[]{1.\textwidth}
    \centering
    \includegraphics[scale=0.6]{figures/legende_modele_compa_lazimpa.png}
    \end{subfigure}
    \caption{Evolution of the accuracy of the three systems for $3$ randomly selected seeds.}
    \label{fig:convergence}
\end{figure*}

We check here the convergence and robustness of our introduced communication system, LazImpa. As a preliminary analysis, we compare the convergence results of: Standard Agents, (Standard Speaker +  Impatient Listener), (Lazy Speaker +  Standard Listener) and LazImpa. In Figure \ref{fig:convergence}, we show the accuracy as a function of the training episodes for $3$ randomly selected seeds. We see that the convergence dynamic is sensitive to the initialization but that in the end, the three systems converge.

\begin{table*}[h]
    \centering
    \begin{tabular}{|c||c|c|c|c|}
    \hline
 & Standard & Lazy Speaker + &  Standard Speaker + & LazImpa \\ 
 & Agents & Standard Listener &  Impatient Listener & \\\hline
$\delta_{stab}$ & $1.16\pm 0.78 \times10^{-3}$  & $1.75 \pm 0.60 \times10^{-3}$ & $9.84 \pm 5.81 \times10^{-5}$  & $9.79 \pm 7.35 \times10^{-5}$ \\ \hline
\end{tabular}
    \caption{Average MSE between the original and smoothed accuracy curve}
    \label{tab:stability}
\end{table*}

Moreover, we observe a gain of stability for the systems with the Impatient Listener. Indeed, as shown in Figure \ref{fig:convergence}, Standard Agents demonstrate a less smooth accuracy curve compared to both (Standard Speaker +  Impatient Listener) and LazImpa. We quantify the stability by introducing a coefficient $\delta_{stab}$ that measures the local variations of the accuracy curves. Formally, we compute the mean square error between the original accuracy curve and the smoothed curve obtained by averaging 10 consecutive score values: 

\begin{equation}
    \delta_{stab} = \frac{1}{n}\sum_{i=1}^{n}(f(i)-\Tilde{f}(i))^{2}
\end{equation}
where $n$ is the total number of episodes, $f(.)$ the accuracy curve (as a function of the number of episode), $\Tilde{f}(i)$ the curve obtained by averaging $f(.)$ over with 11 consecutive episodes centered in $i$. The lower $\delta_{stab}$ is, the smoother the system is .

Results are reported in Table \ref{tab:stability}. $\delta_{stab}$ for systems with Impatient Listener are smaller than the one with Standard Listener confirming the stability of the former. It is important noticing that, contrary to \cite{chaabouni:etal:2019}'s setting where they managed to have more efficient languages at the cost of stable convergence, our new communicative system, on top of leading to efficient languages, has positive impact on the convergence. 

\subsubsection{Complement on randomization test}
\label{appendix:quantif}

To be comparable with \citet{FerrerICancho:etal:2013}, we perform the randomization test with $10^{-5}$ permutations. In the reference article, for a threshold $\alpha$ they introduce two types of p-values:
\begin{itemize}
    \item Left p-value: if left p-value $<\alpha$, the code is characterized by $L_{token}$ significantly smaller than the average weighted message length of any random permutation, corresponding to our notion of \emph{ZLA code}.
    \item Right p-value: if right p-value $<\alpha$, the code is characterized by $L_{token}$ significantly higher than the average weighted message length of any random permutation, corresponding to our notion of \emph{anti-ZLA code}.
\end{itemize}

In the main text, we only report the value of the ZLA significance score $p_{ZLA}$ that is equivalent to \citet{FerrerICancho:etal:2013}'s left p-value. 
However, when also considering right p-value (not shown here), we note for Standard Agents a value smaller than $10^{-5}$ asserting that the system shows a significantly anti-ZLA patterns. 

\subsection{Complements on LazImpa}

\subsubsection{minimal required length by Impatient Listener}
\label{appendix:min_length}

\begin{figure}[h!]
    \centering
    \includegraphics[scale=0.6]{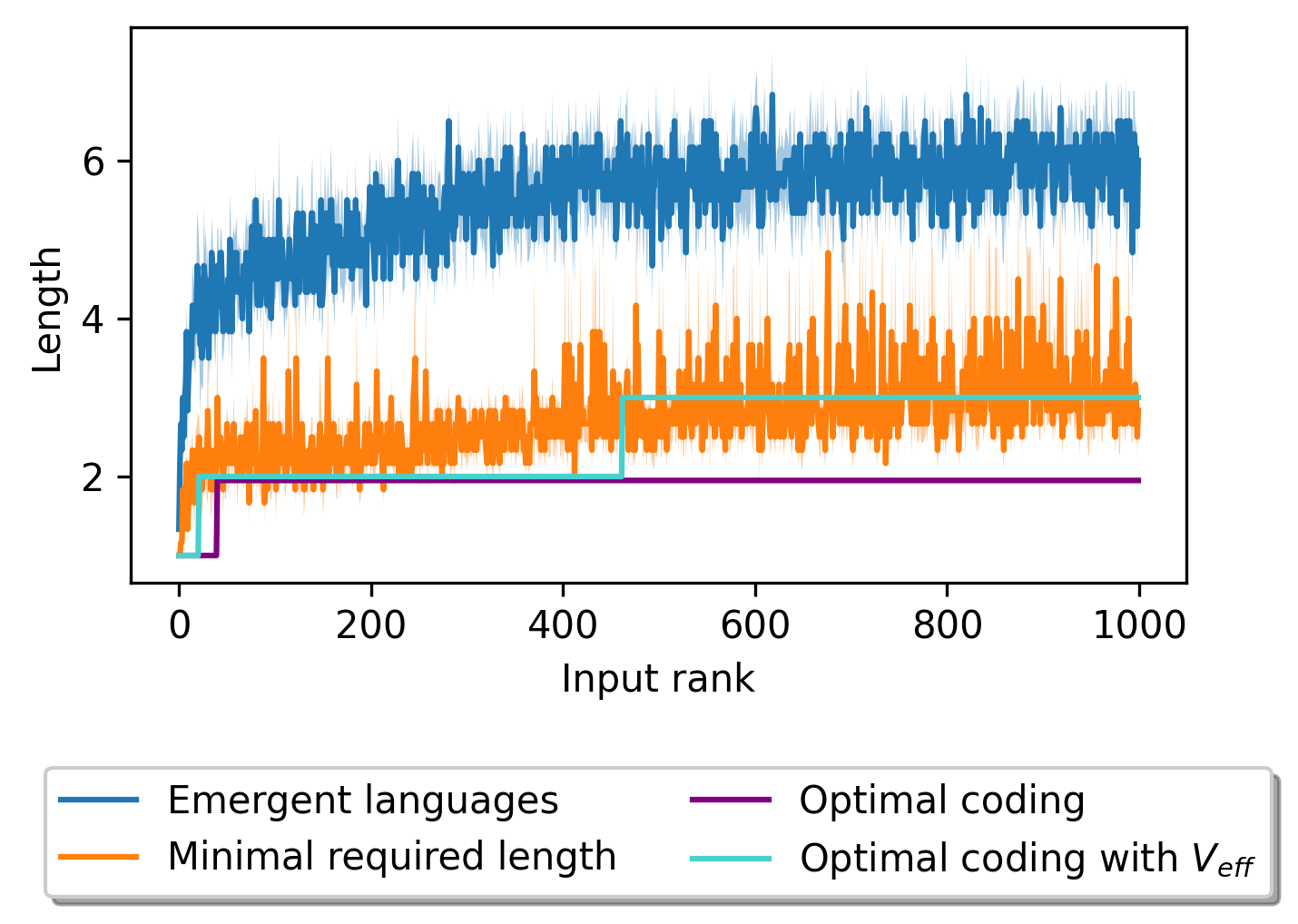}
    \caption{Comparison between the length distribution of the messages and the minimal required length for Impatient Listener to discriminate the messages. The blue curve shows average length distribution function of the inputs frequency ranks. The orange curve represents the average minimal required length by Impatient Listener to decode messages. The purple curve shows the Optimal Coding with the original vocabulary size. The red curve represents the Optimal Coding for the effective vocabulary size $V_{eff}$. Light intervals show 1 standard deviation.}
    \label{fig:min_len_required}
\end{figure}

Thanks to the incremental predictive mechanism of Impatient Listener, it is possible to analyze its intermediate guesses at each reading time. In particular, we are able to spot at which position Impatient Listener is first able to predict the correct output (we verify experimentally that, if Listener finds the correct output at position $i$, it always predicts the right output at position $j>i$). From these intermediate predictions, we define a distribution called `minimal required length' of all the positions at which Impatient Listener is able to first predict the correct output (note that this distribution matches the distribution of the number of informative symbols by message).

We observe that Impatient Listener was often able to find the correct candidate before reading the \verb+EOS+ token. The resulting minimal length is presented in Figure \ref{fig:min_len_required} where we show the length distribution of the messages ranked by input frequency and the actual length required by the Impatient Listener to discriminate the messages. 
We see that the minimal required length by the Impatient Listener is slightly higher than the Optimal Coding. Interestingly, the difference can be partially explained by the use of a skewed distribution of the unigrams across the messages (the Optimal Coding relies on a uniform use of the symbols). Indeed, we compute an effective vocabulary size $V_{eff}$, solution of Equation~\ref{eq:Veff}:

\begin{equation}
    \label{eq:Veff}
    -\sum_{i=1}^{V_{eff}}\frac{1}{V_{eff}}\log\left(\frac{1}{V_{eff}}\right)=\mathcal{H}(\mathcal{U}),
\end{equation}
where $V_{eff}$ is the effective vocabulary size, and $\mathcal{H}(\mathcal{U})$ the entropy of the unigram distribution $\mathcal{U}$ in the emergent communication.

In other words, we search for $V_{eff}$ for which the entropy of a uniform unigram distribution (the left side of Equation~\ref{eq:Veff}) is equal to emergent languages average unigram distribution (the right side of  Equation~\ref{eq:Veff}).

We plot in Figure \ref{fig:min_len_required} a new Optimal Coding with $V_{eff}$ (Optimal Coding with $V_{eff}$). 
The distribution `minimal required length' almost fits the Optimal Coding with this vocabulary size. As shown in Table \ref{tab:len_opt}, the average mean length $L_{type}$ of minimal required length is almost equal to $L_{type}$ of Optimal Coding with $V_{eff}$.

\begin{table*}[h]
    \centering
    \begin{tabular}{|c||c|c|c|}
    \hline
    & Minimal required length & Opt. coding with V & Opt. coding with $V_{eff}$ \\ \hline
    $L_{type}$ & $2.74 \pm 0.08$ & $1.69$ &$ 2.50$ \\
    \hline
    \end{tabular}
    \caption{Comparison of the average length $L_{type}$ of different encoding. `Opt. coding with V' to the Optimal Coding obtained with vocabulary V, `Opt. coding with $V_{eff}$' to the Optimal Coding obtained with vocabulary $V_{eff}$. We also report standard deviation over all the experiments.}
    \label{tab:len_opt}
\end{table*}


\subsubsection{LazImpa robustness to parameters assumptions}
\label{appendix:assumption_analysis}

In this section, we analyze LazImpa robustness to parameters changes. In the main paper, we made two main assumptions: 
\begin{enumerate}
    \item Samples are drawn according to a powerlaw;
    \item 
    $\verb+voc_size+=40$ and $\verb+max_len+=30$.
\end{enumerate}

In the main paper, we demonstrated that LazImpa is able to reach efficient performances with this set of assumptions. We now want to test whether the system is robust to changes of these parameters, i.e.~is LazImpa able to produce efficient and successful codes when inputs are drawn uniformly and/or for different values of $\verb+voc_size+$ ? We report the results of all our experiments in Table \ref{tab:compa_param}. Curves associated to experiments with variations of vocabulary size are shown in Figure \ref{fig:voc_comp}. All these results have been obtained by averaging the results over 3 different seeds by each set of parameters.

\paragraph{Effect of $\texttt{voc\_size}$} : 

\begin{figure}[!h]
    \centering
    \includegraphics[scale=0.6]{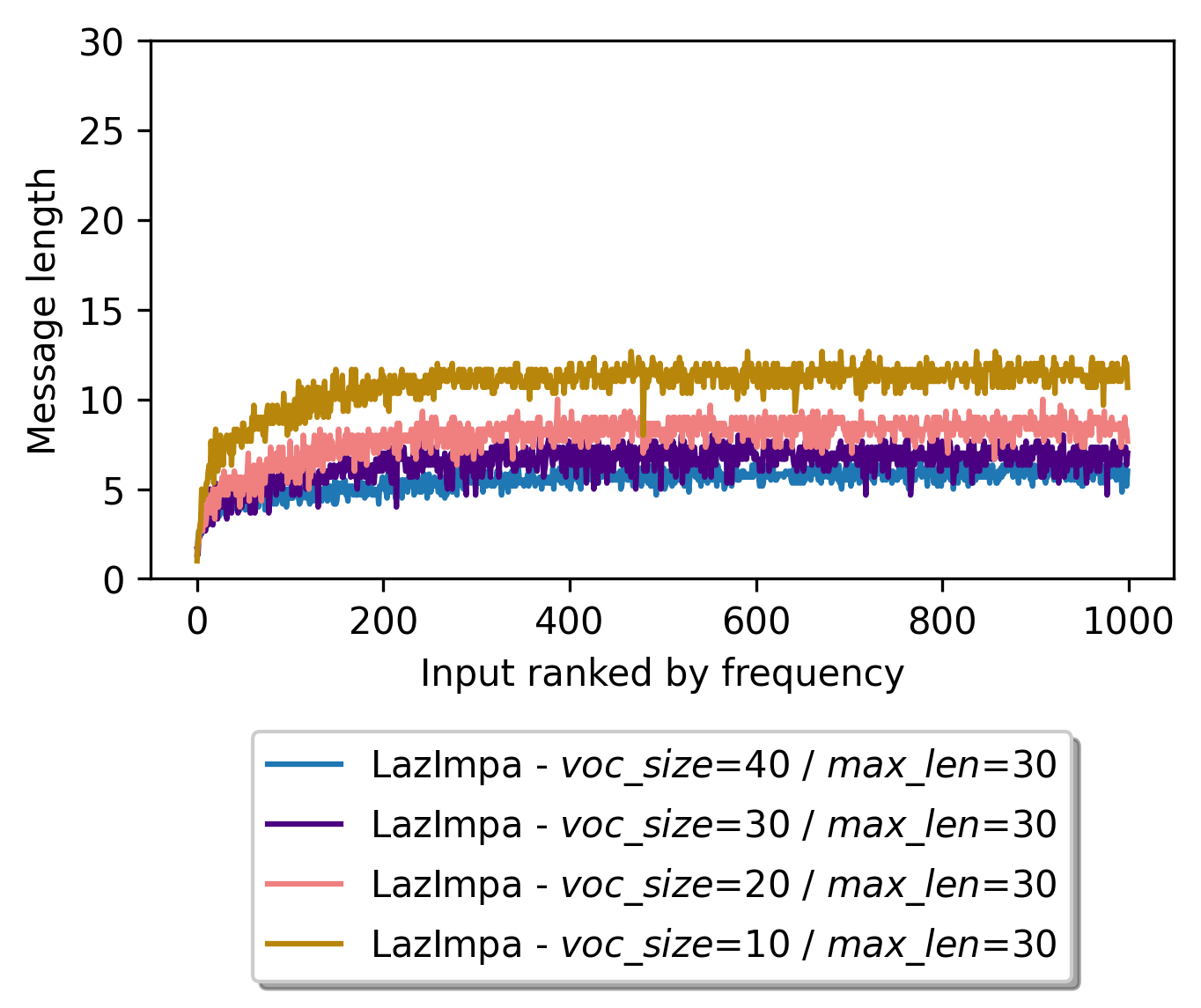}
    \caption{Comparison of LazImpa's average message length for different vocabulary sizes.}
    \label{fig:voc_comp}
\end{figure}
As we can observe in Figure \ref{fig:voc_comp}, emergent codes still respects ZLA for the various tested values of vocabulary size. This is confirmed by the ZLA significance score $p_{ZLA}$ stored in Table \ref{tab:compa_param:voc}. Additionally, we can see a correlation between the size of the vocabulary and the efficiency of the emergent code: the emergent code is more efficient for large sizes of vocabulary. Indeed, we observe that $L_{type}$, $L_{token}$ and $L_{eff}$ are increasing functions of the vocabulary size. This is expected as the number of messages of a given length increases with the vocabulary size. Thus, the set of `short' messages is higher for a large vocabulary size. Naturally, the same trend is observed with Optimal Coding. Moreover, we note a decrease  of $\rho_{inf}$ as a function of \verb+voc_size+ for the LazImpa system, suggesting that the smaller the vocabulary size is the more noninformative positions are used.

\paragraph{Effect of $\texttt{max\_len}$:} We can note in Table \ref{tab:compa_param:max_len} that LazImpa is even closer to Optimal Coding when setting $\texttt{max\_len}=20$. $L_{type}$, $L_{token}$ and $L_{eff}$ are slightly smaller compared to experiments with $\texttt{max\_len}=30$. Thus, agents regularization seems to be easier when setting smaller values of $\texttt{max\_len}$. Nevertheless, the results are very close. In particular, we can note that information density values $\rho_{inf}$ are very similar suggesting that sub-optimality issues are independent of the parameter $\texttt{max\_len}$. Note that we only explore two values of $\texttt{max\_len}$ in Table \ref{tab:compa_param:max_len} because small and large values of $\texttt{max\_len}$ lead respectively to a small and large message space and thus optimization issues (H-parameters tuning is required to favor respectively exploration and exploitation).

\paragraph{Effect of input distribution: } As we observe in Table \ref{tab:compa_param:distrib}, LazImpa's performances are quite similar when dealing with inputs drawn from a uniform or a powerlaw distribution. In particular, with a uniform distribution, we observe a gain of efficiency for $L_{type}$ and a loss of efficiency for $L_{token}$ while $L_{eff}$ is almost unchanged. All these results are expected. Equal $L_{eff}$ means that Impatient Listener relies on the same number of symbols on average. In the main paper, we have shown that $L_{eff}$ is mostly influenced by the entropy of the unigram distribution. Since, there is no change of $\texttt{voc\_size}$, we do not expect major changes of entropy and thus no change for $L_{eff}$. Then, the difference of $L_{token}$ and $L_{type}$ is explained by the reduction step. For uniformly drawn inputs, the regularization is uniformly applied on the inputs ; for inputs drawn from a powerlaw, the regularization mostly focuses on the most frequent inputs because they have larger weights in the loss. Consequently, we expect a lower $L_{token}$ when experimenting with a powerlaw distribution, compared to the uniform setting, but a larger $L_{type}$. Eventually, we observe a significant gain of information density $\rho_{inf}$ for LazImpa with a uniform distribution. This is mainly explained by $\rho_{inf}$ computation that takes into account message lengths without involving their frequency. 

As a remark, let's precise that we do not explore a larger set of non-uniform input distributions. In theory, the shape of the length distribution should not be impacted by the input distribution because the optimization problem is only dependent of the frequency ranks (mapping of the shortest messages to the most frequent inputs).

\begin{table*}[h]
    \centering
    \captionsetup[subtable]{position=b}
    \begin{subtable}{1\linewidth}
    \begin{adjustbox}{max width=\textwidth}
        \begin{tabular}{|c||c||c|c|c|c|c|}
        \hline
        
             $\verb+voc_size+$ & System & $L_{type}$ & $L_{token}$ & $p_{ZLA}$ & $L_{eff}$ & $\rho_{inf}$  \\
            \hline
            \hline
             40 & LazImpa & $5.49 \pm 0.67$ & $3.78 \pm 0.34$ & $<10^{-5}$* & $2.67 \pm 0.07$ & $0.60 \pm 0.07$  \\ \cline{2-7}
              & Optimal Coding & $2.96$ & $2.29$ & $<10^{-5}$* & $1.96$ & $1$  \\ 
             
             
             \hline 
             \hline
             30 & LazImpa & $6.49 \pm 1.20$ & $4.14 \pm 0.43$ & $<10^{-5}$* & $2.71 \pm 0.22$ & $0.53 \pm 0.07$ \\ \cline{2-7}
             & Optimal Coding & $3.09$ & $2.35$ & $<10^{-5}$* & $2.09$ & $1.$ \\
             
             \hline 
             \hline
             20 & LazImpa & $7.91 \pm 0.71$ & $4.80 \pm 0.30$ & $<10^{-5}$* & $2.98 \pm 0.07 $ & $0.45 \pm 0.04$  \\ \cline{2-7} 
             & Optimal Coding & $3.59$ & $2.51$ & $<10^{-5}$* & $2.59$ & $1.$  \\ 
             
             \hline 
             \hline
             10 & LazImpa & $10.82 \pm 0.28$ & $6.54 \pm 0.06$ & $<10^{-5}$* & $3.87 \pm 0.10$ & $0.40 \pm 0.005$  \\ \cline{2-7}
              & Optimal Coding & $4.08$ & $2.82$ & $<10^{-5}$* & $3.08$ & $1.$  \\ 
             
             \hline 
        \end{tabular}
        \end{adjustbox}
        \caption{Variations of vocabulary size $\texttt{voc\_size}$. By default, the input distribution is a powerlaw and  $\texttt{max\_len}=30$. \label{tab:compa_param:voc}}
    \end{subtable}
    \bigbreak
    \begin{subtable}{1\linewidth}
    \centering
    \begin{adjustbox}{max width=\textwidth}
        \begin{tabular}{|c|||c||c|c|c|c|c|}
        \hline
        
              $\texttt{max\_len}$ & System & $L_{type}$ & $L_{token}$ & $p_{ZLA}$ & $L_{eff}$ & $\rho_{inf}$  \\
            \hline
            \hline
         30 & LazImpa & $5.49 \pm 0.67$ & $3.78 \pm 0.34$ & $<10^{-5}$* & $2.67 \pm 0.07$ & $0.60 \pm 0.07$  \\ \cline{2-7}
              & Optimal Coding & $2.96$ & $2.29$ & $<10^{-5}$* & $1.96$ & $1$  \\ 
             
             \hline 
             \hline
             20 & LazImpa & $4.36 \pm 0.11$ & $3.12 \pm 0.06$ & $<10^{-5}$* & $2.40 \pm 0.08$ & $0.55 \pm 0.01$ \\  \cline{2-7}
               & Optimal Coding & $2.96$ & $2.29$ & $<10^{-5}$* & $1.96$ & $1$ \\  
             
             \hline 
    \end{tabular}
    \end{adjustbox}
    \caption{Variations of maximum length $\texttt{max\_len}$. By default, the input distribution is a powerlaw and $\texttt{voc\_size}=40$. \label{tab:compa_param:max_len}}
    \end{subtable}
    \bigbreak
    \begin{subtable}{1\linewidth}
    \centering
    \begin{adjustbox}{max width=\textwidth}
        \begin{tabular}{|c|||c||c|c|c|c|c|}
        \hline
        
              Distribution & System & $L_{type}$ & $L_{token}$ & $p_{ZLA}$ & $L_{eff}$ & $\rho_{inf}$  \\
            \hline
            \hline
             powerlaw & LazImpa & $5.49 \pm 0.67$ & $3.78 \pm 0.34$ & $<10^{-5}$* & $2.67 \pm 0.07$ & $0.60 \pm 0.07$  \\ \cline{2-7}
              & Optimal Coding & $2.96$ & $2.29$ & $<10^{-5}$* & $1.96$ & $1$  \\ 
             
             \hline 
             \hline
             uniform & LazImpa & $4.27 \pm 0.37$ & $4.27 \pm 0.37$ & / & $2.53 \pm 0.09$ & $0.81 \pm 0.08$ \\  \cline{2-7}
               & Optimal Coding & $2.96$ & $2.96$ & / & $1.96$ & $1$ \\  
             
             \hline 
    
    \end{tabular}
    \end{adjustbox}
    \caption{Variations of input distribution. By default: $\texttt{voc\_size}=40$, $\texttt{max\_len}=30$. \label{tab:compa_param:distrib}}
    \end{subtable}
    \caption{Efficiency analysis of LazImpa and Optimal Coding for different set of parameters. $L_{type}$ is the mean message length, $L_{token}$ is the mean weighted message length, $p_{ZLA}$ the ZLA significance score, $L_{eff}$ the effective length and $\rho_{inf}$ the information density. `/' indicates that the metric is not relevant. For $p_{ZLA}$, `*' indicates that the p-value is significant ($<0.001$).}
    \label{tab:compa_param}
\end{table*}

\subsubsection{Statistical comparison between LazImpa and natural languages}
\label{appendix:stat_NL}

\begin{figure*}[h]
    \centering
    \includegraphics[scale=0.4]{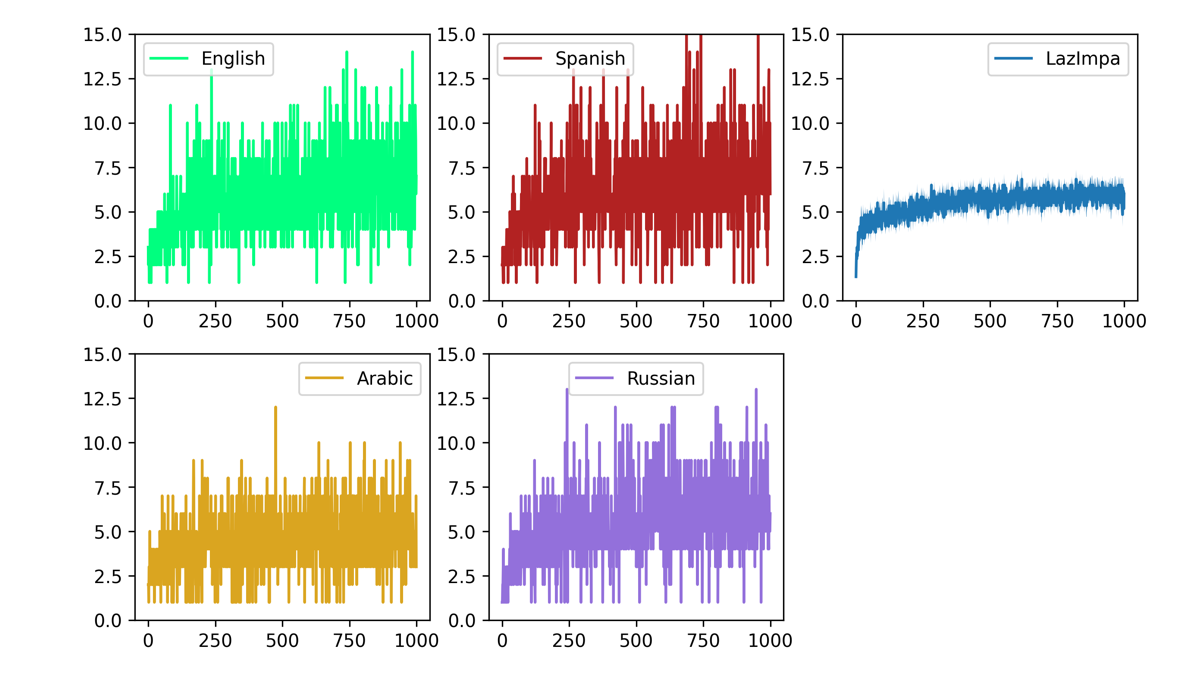}
    \caption{Comparison of the message length as a function of input frequency rank for LazImpa and natural languages.}
    \label{fig:compa_NL_entier}
\end{figure*}

Figure \ref{fig:compa_NL_entier} shows the words length as a function of their frequency for both natural languages and the emergent language. This figure completes our comparison made in the main paper between LazImpa and natural languages where curves were smoothed. Here we show the raw natural languages distribution. The additional observation that we can make is that the variance of the words length is larger for the natural languages.

\end{document}